\pdfoutput=1

\documentclass[11pt]{article}

\usepackage[]{latex/acl}

\usepackage{times}
\usepackage{latexsym}

\usepackage{xcolor}
\definecolor{graytext}{gray}{0.5} 

\usepackage[T1]{fontenc}

\usepackage[utf8]{inputenc}

\usepackage{microtype}

\usepackage{inconsolata}

\usepackage{graphicx}

\usepackage{xspace}
\usepackage{tabularray}
\usepackage{amsmath}
\usepackage{algorithm}
\usepackage{algorithmic}
\usepackage{multirow}
\usepackage{paralist}
\usepackage{xcolor}
\usepackage{booktabs}
\usepackage{tabularx}
\usepackage{booktabs}

\newcommand{\remove}[1]{}

\newcommand{\base}{\ensuremath{\mathcal{B}}\xspace}
\newcommand{\targ}{\ensuremath{\mathcal{T}}\xspace}
\newcommand{\targprime}{\ensuremath{\mathcal{T}'}\xspace}

\title{ParallelPARC: A Scalable Pipeline for Generating \\ Natural-Language Analogies}

\author{Oren Sultan, Yonatan Bitton, Ron Yosef, Dafna Shahaf \\
The Hebrew University of Jerusalem \\
\texttt{\{oren.sultan, yonatan.bitton, ron.yosef, dshahaf\}@cs.huji.ac.il}}

\begin{document}
\maketitle
\begin{abstract}



Analogy-making is central to human cognition, allowing us to adapt to novel situations -- an ability that current AI systems still lack.
%
%
%
Most analogy datasets today focus on simple analogies (e.g., word analogies); datasets including complex types of analogies are typically manually curated and very small. We believe that this holds back progress in computational analogy.

In this work, we design a data generation \emph{pipeline}, \emph{ParallelPARC (Parallel Paragraph Creator)} leveraging state-of-the-art Large Language Models (LLMs) to create complex, paragraph-based analogies, as well as distractors, both simple and challenging. 
We demonstrate our pipeline and create \emph{ProPara-Logy}, a dataset of analogies between \emph{scientific processes}. We publish a \emph{gold-set}, validated by humans, and a \emph{silver-set}, generated automatically. We test LLMs' and humans' analogy recognition in \emph{binary} and \emph{multiple-choice} settings, and found that humans outperform the best models ($\sim$13\% gap) after a light supervision.
We demonstrate that our \emph{silver-set} is useful for training models. Lastly, we show \emph{challenging distractors} confuse LLMs, but not humans.
We hope our pipeline will encourage research in this emerging field.




\end{abstract}

\section{Introduction}

\begin{figure*}[t]
\includegraphics[width=.9\textwidth]{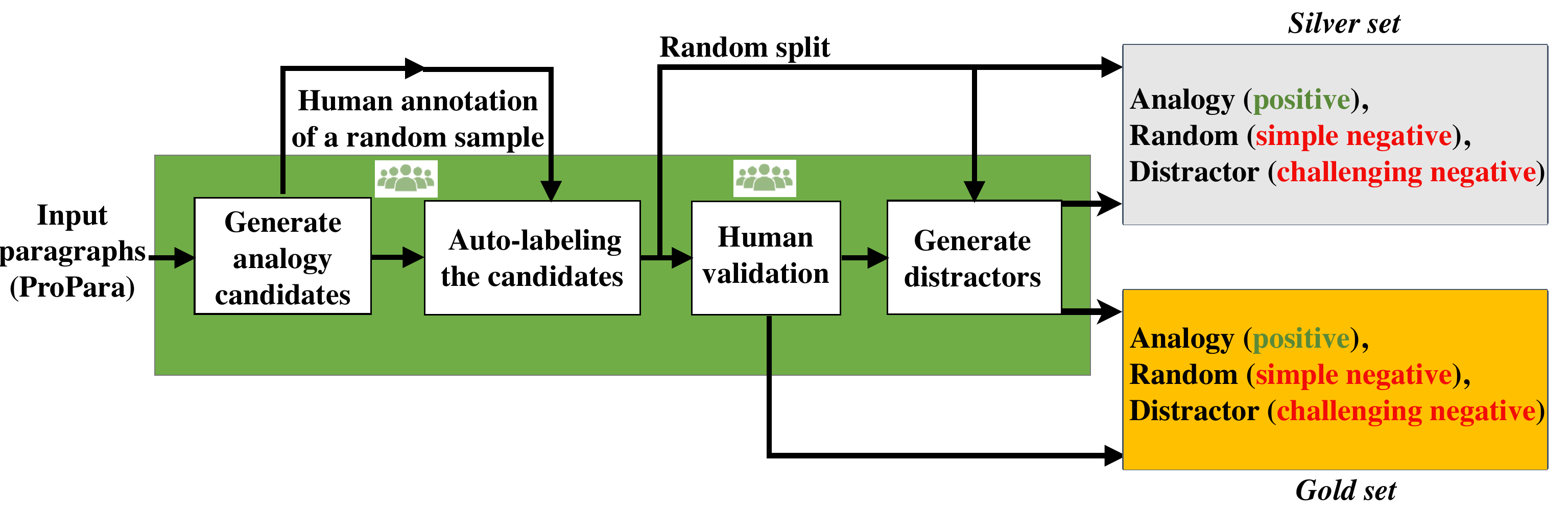}
\centering
\caption{Our data generation pipeline. We {generate analogy candidates}, then collect {human annotations on a random sample} to be used as few-shot for an auto-labeling model. We {run the model} to label candidates at scale.
We randomly split the data into \emph{silver-set} and \emph{gold-set}, which is {validated by humans}. 
In addition to \emph{positives} (analogies), we include random target paragraphs ({simple negatives}), 
and {generate distractors} ({challenging negatives}).
}
\label{fig:data_generation_pipeline}
\end{figure*}

Analogy-making is a central to human cognition. It allows us to abstract information and understand novel situations in terms of familiar ones \citep{minsky1988society,hofstadter2013surfaces,holyoak1984analogical}, abilities that are still lacking in current AI systems.
Research suggests that these abilities are essential for robust AI  that can effectively generalize and adapt to diverse domains \citep{Mitchell2021AbstractionAA}.

According to Gentner’s Structure Mapping Theory (SMT) \cite{GENTNER1983155}, analogy is a \emph{mapping} from entities in {base \base} to entities in {target \targ}, relying on \emph{relational similarity}, not \emph{object attributes}. For example, in the analogy between an {electrical circuit} and a {water pump}, there is a mapping between \emph{electrons → water}, \emph{wire → pipe}. While object attributes are different ({water} is liquid, {electrons} are not), the \emph{relations} are similar ({electrons} move through {wires} like {water} flows in {pipes}).

Despite the importance of analogy, relatively few analogy resources exist today. Most resources mainly focus on \emph{word-analogies} (``{A}:{B} is like {C}:{D}''). We argue that this setting is too simplistic, often boiling down to a single relation (``PartOf'', conjugation); in the real world, analogies are often complex, involving multiple entities and intricate relations between them. Real-world analogies are often described in natural language, adding to the complexity of the problem.  
%
%
A very recent work employed LLMs to generate analogies at scale between 2-sentence snippets ($\sim$20 tokens) \cite{jiayang2023storyanalogy}. However, resources of more complex analogies (e.g., full paragraphs) are few and \emph{sparse} (18 samples max).
{We believe this lack of data hinders progress in computational analogy; in the past, high-quality datasets have led to a burst of novel research (e.g., ImageNet \cite{5206848}).}

In this work, we design a pipeline, \emph{ParallelPARC (Parallel Paragraph Creator)} to scale up the process of generating analogies between paragraphs (see Figure \ref{fig:data_generation_pipeline}), leveraging recent progress in LLMs. 
We release  a \emph{gold-set}, validated by humans, and a \emph{silver-set}, which is automatically generated. 

Coming up with non-trivial negative examples (non-analogous paragraphs) is a challenging task. Our pipeline generates, in addition to  {positives}  (analogies), both
\emph{simple negatives} (random paragraphs) and \emph{challenging negatives} (distractors).

\begin{figure*}[t]
\includegraphics[width=\textwidth]{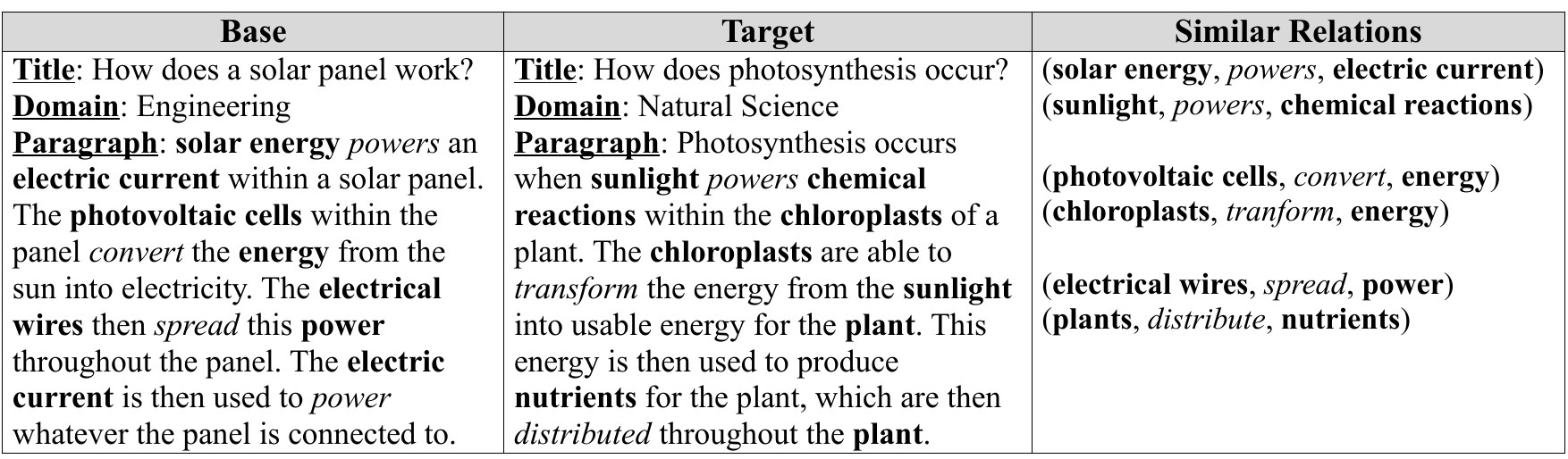}
\caption{An example of an \emph{analogous} sample {from our dataset (generated by our pipeline)}. Two scientific processes, base and target, are described via a title, a domain, and a paragraph of natural-language text. A sample also includes \emph{similar relations}, hinting at why the processes are analogous. 
}
\label{fig:dataset_example}
\end{figure*}

To demonstrate our pipeline, we create \emph{ProPara-Logy}, a dataset of paragraphs describing \emph{scientific processes} across various  domains, meant for studying analogical reasoning. 
A sample in our data includes two processes, each 
described via a title (``How does a solar panel work?''), a domain (``Engineering''), and a full paragraph. 
In addition, the data includes \emph{similar relations} between the two processes, which is a core part in understanding why they could be analogous (See Figure~\ref{fig:dataset_example}).

We  evaluate LLMs and humans on  \emph{binary} and \emph{multiple-choice} analogical reasoning tasks on  \emph{ProPara-Logy}. We found that humans outperform the best models ($\sim$13\% gap) after a light supervision. We show  the automatically-generated \emph{silver-set} is useful for training models, and can significantly improve their performance.  Finally, we demonstrate the distractors significantly reduce the performance of LLMs,  but not  of humans.

\textbf{Our main contributions are:} 

{
\begin{compactitem}
 \item We develop a novel data pipeline to create complex, paragraph-based analogies. 
 \item We demonstrate our pipeline and create the \emph{ProPara-Logy} benchmark, a dataset for analogical reasoning over paragraphs describing processes in science. Our dataset is orders of magnitude larger than previous work, and could easily be expanded. 
 
 \item Beyond the analogous paragraphs (positives), our dataset includes both simple and challenging distractors (negatives). It also includes useful information about the analogies, such as relations shared between the paragraphs. 

 \item We use \emph{ProPara-Logy} to evaluate humans and LLMs on our proposed analogical reasoning tasks, both in zero-shot and guided settings.
 
  \item We release data and code at \url{ https://github.com/orensul/ParallelPARC}.
\end{compactitem}
}

\section{Existing Analogy Datasets}

We now survey available analogy resources.

\noindent\textbf{Word analogies.}
Many analogy resources focus on \emph{word analogies} (``{A}:{B} is like {C}:{D}'') \cite{jurgens-etal-2012-semeval, Popov2017TheRL, KMIECIK201925, Rogers2016AnalogybasedDO, Czinczoll2022ScientificAC}. Such analogies are widely used in entrance tests like the SAT in the US or NCEE in China.

This area has gained popularity in the NLP community after \citet{Mikolov2013EfficientEO} show that word embeddings can model some relational similarities in terms of word vector offsets. This method can find analogies relying on certain simple types of relations, but struggles with complex relations \cite{linzen-2016-issues, schluter-2018-word, ushio2021bert}.
More recently, several studies explored the use of LLMs in generating word analogies \cite{bhavya-etal-2022-analogy, Yuan2023BeneathSS, Yuan2023ANALOGYKBUA}.


In addition to the word analogy itself (A, B, C and D), some resources include extra information, such as explanations \cite{chen2022kar}. Other resources include multiple correct options, either close analogies (C, D are similar to A, B) or far (C, D are from a different domain than A, B) \cite{PMID:19383937}. Some resources include wrong answers, but often quite simple (e.g., random words).


\noindent\textbf{Visual analogies.}
This is the visual equivalent of word analogies (where A, B, C, D are images).
There have been multiple attempts to represent \emph{transformations} between pairs of images \cite{reed2015deep,radford2015unsupervised, tewel2021zero}, typically stylistic or geometric, and several resources published~\cite{sadeghi2015visalogy,Bitton2022VASRVA}.
As in word analogies, generating wrong answers is challenging. They are often created by either using random images or images that contain elements of the correct answer but exclude another element that is crucial for the analogy.


A different kind of resource is ARC~\cite{Chollet2019OnTM}, where test-takers  have to discern rules from \emph{pixel grids} to deduce the correct output grid. 


\noindent\textbf{Paragraph-level analogies.} Very recently, \citet{jiayang2023storyanalogy} created a dataset of 24K story pairs. However, the pairs are short snippets (2 sentences, $\sim$20 tokens), and well-aligned, making the setting overly simplistic. Moreover, their work does not assess directly whether a pair is analogous. 
There are few resources of analogies between \emph{full paragraphs}, most notably stories from cognitive-psychology literature \cite{gentner1993roles, wharton1994below, clement1991systematicity}. These datasets are manually curated and very small (18 samples max), rendering them inadequate for training models. Furthermore, the stories have a near-identical structure (``Mr. Newton was the manager of a company that made razors"/``Mr. Boyce was director of manufacturing shaving knives...''), again making the setting non-realistic.

Notably, the dataset of \citet{gentner1993roles} also includes {false analogy stories}, which are 
similar to the base paragraph in terms of {first-order relations}, but dissimilar in {higher-order relations} (relations between the first-order relations). 
\citet{jiayang2023storyanalogy} includes simple (random) negatives and hard negatives (snippets with similar entities).

A recent work focused on finding analogies between {paragraphs describing processes} \cite{sultan-shahaf-2022-life}. Their method ranks pairs of paragraphs from a dataset, such that analogous pairs rank high. However, this is a noisy resource, as many non-analogies  rank high, and many of the identified analogies are from very close topics.


\label{sec:existing_datasets}

\section{Dataset Generation}

Our goal is to develop a \emph{pipeline} for generating high-quality data that could drive forward research efforts in computational analogy.  Figure \ref{fig:dataset_example} illustrates the format of data generated by {our pipeline. A records contains two processes, base \base and target \targ. Each process is described via a title (``How does a solar panel work?''), a domain (``Engineering''), and a full paragraph.} 

In addition to expressive natural-language paragraphs, the data also includes  \emph{similar relations} between the two processes, which is a core element in identifying analogies (Figure~\ref{fig:dataset_example}, right). 

Figure~\ref{fig:dataset_example} shows a \emph{positive example} (analogy). 
In addition to {positives}, our pipeline generates \emph{simple negatives} and \emph{challenging distractors}, designed to fool both humans and models. 

The pipeline (see Figure~\ref{fig:data_generation_pipeline}) begins by using LLM for generating \textbf{analogy candidates} -- paragraphs (and relations) that {potentially} describe analogous processes across diverse domains in science {(\S\ref{subsec:analogyCandidatesGeneration})}.
Then, we use {\bf human annotators} to label a random sample of the candidates {(\S\ref{subsec:humanInitialAnnotation})}, and  use the annotated data as a few-shot for \textbf{automatic labeling} of candidates   {(\S\ref{subsec:analogyFiltering})}. Then, we filter the data based on the automatic labels, and randomly split the filtered data into two disjoint sets: our \emph{gold-set}, further \textbf{validated by humans}, and our  \emph{silver-set}, which is not {(\S\ref{subsec:humanAnnotation})}.
Finally, we employ an LLM to generate challenging \textbf{distractors} {(\S\ref{subsec:distractors})}.

\label{sec:dataset}

\subsection{Analogy Candidates Generation}
\label{subsec:analogyCandidatesGeneration}

Our goal in this section is to generate analogy candidates from diverse scientific domains. 

We employed GPT-3.5 (text-davinci-003)\footnote{We have chosen GPT-3.5 after experimenting with several newer models, and finding that it delivers high-quality results at a very reasonable cost.} \cite{Brown2020LanguageMA} (see implementation details in  Appendix~\ref{subsec:appendix_analogy_candidates_generation}).
We first na\"{\i}vely tried to  ask GPT repetitively for two analogous scientific processes (with no additional constraints or guidance). We found that GPT (1) tends to repeat itself, and (2) often creates analogies revolving around extremely similar topics.


To solve the problem of \emph{repetitiveness}, we seeded GPT with \base instead of asking for generating both \base and \targ.
We used the ProPara dataset \citep{proparNaacl2018} of English paragraphs describing scientific processes, taking 390 titles from its training set. 
To solve the problem of \emph{similar topics}, we tried to explicitly diversify the target paragraphs by asking for analogies in specific fields (e.g., zoology), but often no analogies were found. Ultimately, we selected several \emph{broad} domains: \emph{Engineering}, \emph{Natural Science}, \emph{Social Science} and \emph{Biomedical and Health Science}. 
This provided a balance between diversity and specificity, and also allowed us to control the distribution of target domains. 


We first tried using a single prompt for generating analogies. However, that led to paragraphs that were mostly identical to the input paragraph except for nouns (``The sediment is deposited again in a new place''/``Money is deposited again in a new place''), and artificially sounding sentences (``Money travels through the economy'').  

As noted earlier, analogy is often defined as a system of similar relations \cite{GENTNER1983155}. Thus, we decided to use relations as a stepping stone towards generating analogies; we developed two \emph{separate prompts}, one for finding an analogous subject and identify similar relations, and another for taking the subject and relations and turning them into paragraphs in natural language (see Appendix~\ref{subsec:appendix_analogy_candidates_generation} Figures~\ref{tab:analogy_candidates_generation_prompt1}, \ref{tab:analogy_candidates_generation_prompt2}). This approach has proven to be effective in practice. 
{We experimented with one-shot and few-shot settings, and chose the one-shot prompt, which was more cost-effective.}

We include relations in our data in addition to the paragraphs, subjects, and domains (Figure~\ref{fig:dataset_example}, right). We believe they are can also serve as potential \emph{explanations}, highlighting the structural similarity between base and target paragraphs. 
%

For each paragraph in ProPara, we generate 3 analogy candidates in 4 broad domains, resulting in 4680 samples. We filter out samples with less than 3 similar relations (less likely to be analogies), leaving us with 4288 candidates.

\subsection{Human Annotation Task}
\label{subsec:humanInitialAnnotation}

\begin{figure*}[!t]
    \centering
    \resizebox{\textwidth}{!}{%
        \fbox{\parbox[c]{6cm}{
            \textbf{{Base}}:  \\ \textbf{How do bats use echolocation? \\ (Natural Sciences)} \\
            Bats use echolocation to navigate and find food. \textbf{They emit high frequency sound waves} that bounce off of objects in their environment.
            \\ The bats then \textbf{receive the echoes} and \textbf{interpret the information} to locate their prey and navigate their surroundings. Submarines interpret the echo to determine the distance and size of the object.
        }}
        \fbox{\parbox[c]{5.5cm}{%
            \textbf{{Target (Analogy)}}: \\ \textbf{How do submarines use sonar? (Engineering)} \\
            Submarines use sonar technology to detect objects in the water. \\ 
            \textbf{They emit sound waves}, which travel through the water and bounce off the objects. \\ \textbf{The sound waves are then received back as an echo}. \textbf{Submarines interpret the echo} to determine the distance and size of the object.
            
        }}
        \fbox{\parbox[c]{5.6cm}{%
            \textbf{{Target (Distractor)}}: \\ \textbf{How do submarines use sonar? (Engineering)} \\
            \textbf{Submarines interpret the echo} to determine the distance and size of the object. \textbf{After interpreting the echo, they emit sound waves}, which travel through the water and bounce off the objects. \textbf{These sound waves are then received back as an echo}. Finally, submarines use sonar technology to detect objects in the water.
            
        }}
    }
    \caption{
        {An example of the distractor creation process. On the left is the \textbf{{Base}} paragraph (about bats using echolocation). In the middle, a \textbf{{Target}} paragraph, which is analogous to the base paragraph. On the right is a \textbf{{Target (Distractor)}} paragraph, generated from the middle paragraph by switching the order of events: The emission of sound waves, followed by their reception as an echo, and submarines interpret the received echo. In the \textbf{{Target (Distractor)}}, the order is reversed, altering the cause-and-effect relations from the true analogy.}
    }
    \label{fig:distractor_change_order_example}
\end{figure*}

In the previous section we generated analogy candidates. We now annotate a small portion of this data. 
Our goal is two-fold: (1) to estimate the proportion of analogies in the data, as well as identify issues with the generation process, and (2) to use the annotated data to train models.
 
We hired Amazon Mechanical Turk (AMT) workers who passed a rigorous qualification task.
Workers received two paragraphs, base \base and target \targ, corresponding subjects, domains, and the similar relations generated by the LLM.
The task is to determine whether the paragraphs are analogous \emph{and} the similar relations are correct. 
If they are, the worker needs to select between close analogy (close topic, similar entities) or far analogy (unrelated topics).
If there is an issue with the analogy or the relations, the worker marks it ``for further inspection'', along with a reason: {dissimilar relations}, {misinformation}, {cyclic vs. non-cyclic process}, or other (with a free-text explanation).

Note that two processes may be deemed analogous or not depending on the annotator's abstraction, which is affected by their domain knowledge.
To ameliorate this, we explicitly instructed annotators to focus on relational similarity, between relations \emph{as they are expressed in the texts}, and not take domain knowledge into account.

Three workers labeled each sample, for a reward of \$0.5 per sample. See Appendix~\ref{subsec:appendix_human_annotation} for more details about the annotation process.

\subsection{Automatic Filtering and Labeling}
\label{subsec:analogyFiltering}


Based on the annotations in Section \ref{subsec:humanInitialAnnotation}, we estimate analogies to be less than 30\% of the dataset.

Next, we decided to use part of our annotated data as few-shot examples for our \emph{filtering model}. The goal is two-fold: (1) As the annotation process is long and costly, it could  identify the most probable analogous candidates to show our annotators. (2) If the model performance matches humans, we could replace the human-in-the-loop and achieve a {\bf fully automated} pipeline. 

This task is complex, and thus we use GPT-4 \cite{openai2023gpt}, a state-of-the-art LLM (parameters in Appendix~\ref{subsec:appendix_filtering}).
We input randomly selected annotated candidates (30 examples, maximum allowed tokens) into GPT, comprising two paragraphs, their subjects, similar relations, and a label indicating how many workers labeled it as an analogy (0-3).
See Appendix~\ref{subsec:appendix_filtering} for the prompt.

Following the in-context learning phase, we run the model on our unlabeled analogy candidates.

\subsection{Human Validation}
\label{subsec:humanAnnotation}
Our goal in this paper is to demonstrate how our pipeline can be used for creating datasets. 
We consider two types of datasets: a \emph{silver-set}, automatically labeled, and a \emph{gold-set}, validated by humans.  

Thus, we returned to the task from Section \ref{subsec:humanInitialAnnotation}. We show annotators both the most likely analogous candidates, as predicted by the model, but also the least likely candidates.  This allows us to evaluate the filtering model where it is most certain. It also balances the data for the annotators. 
%
In addition, it is surprisingly hard to come up with hard negative examples. We believe that the least likely candidates (according to the model) hold the potential to be useful in future research (see Section \ref{subsec:dataset_summary}).

{Identifying analogies is a complex task. Therefore, in addition to the thorough qualification phase (Section~\ref{subsec:humanInitialAnnotation}), we also consistently monitored and provided clarifications to the annotators.
To further ensure quality for our \emph{gold-set},  we chose a strict setting: a sample is positive only if \emph{all three} annotators agree it is an analogy.}
For our proof-of-concept, we wanted a \emph{gold-set} with at least 300 positives. We randomly gave annotators small batches to label until reaching 310 positives. 
%
Annotators labeled 828 instances (not including the 130 from Section \ref{subsec:humanInitialAnnotation}), for a total cost of \$1,804. 

Our annotators' agreement is 78.6\%, where random chance is 25\% (\% of perfect agreement).


\noindent\textbf{Filtering model evaluation.}
We also use the annotated data to evaluate the filtering model. We compare its predictions to workers' majority vote. 
Our model achieves an accuracy of 85.1\%, f1-score of 83.4\%.
Importantly, it reaches 79.5\% precision when predicting high likelihood of an analogy, which is significantly higher than the 30\% base rate, and 90\% precision when predicting low likelihood. 
%
These results show our model reliably replicates annotators on the high-confidence samples, rendering our approach \emph{scalable}. 
Consequently, we release a \emph{silver-set}, generated by applying the filtering model on the remaining candidates. This data could also be useful for training models (Section \ref{sec:experiments}).



\subsection{Distractors Generation}
\label{subsec:distractors}

In addition to the 310 analogies in our \emph{gold-set}, we create  \emph{simple} negatives from random ProPara paragraphs on different subjects\footnote{\label{random_paragraph_footnote}
We estimate two ProPara paragraphs on different subjects are analogous in $\sim$1\%, based on \citet{sultan-shahaf-2022-life}.} as \targ. However, those are quite easy to tell apart from analogies; thus, we now focus on creating \emph{challenging} negatives.


While many types of distractors are possible, we are inspired by the ideas in \citet{gentner1993roles}.
There, story pairs match or not match at three levels: attributes, first-order  and higher-order relations. We focus on the most {\it complex and challenging} setting -- stories matching only in first-order relations. 
We leave other dimensions for future work.


\noindent\textbf{Formulation.}
Let \base and \targ be two analogous paragraphs. 
The intuition is to create distractor \targprime that keeps first-order relations of \targ (Figure~\ref{fig:dataset_example}, right) but changes the higher-order relations -- i.e., relations between first-order relations, such as cause and effect, or temporal dependencies between events. 
%
%
To create \targprime, we find two \emph{dependent events} in \targ such that one must precede the other, and switch their order. 
See Figure~\ref{fig:distractor_change_order_example} for an example, generated by our method: the relations are the same, but the submarines interpret the echo \emph{before} emitting sounds returning as echos.
See Appendix~\ref{subsec:appendix_distractors} for details.

\noindent\textbf{Generation.}
We use GPT-4 to automatically generate distractors with two separate prompts: (1) finding and replacing two dependent events, and (2) writing a \emph{coherent} \targprime. For the first task, we use one-shot. We ask GPT-4 to output a list of the events in \targ according to their order in time, and then replace two dependent events, along with an explanation. For the second task, we use few-shot. The input is an order of events and the output is a \emph{coherent} paragraph. See details in Appendix~\ref{subsec:appendix_distractors}. 

\noindent\textbf{Evaluation.}
We now evaluate the generated distractors. A {correct distractor} should switch two dependent events, with a paragraph that is coherent and consistent with the new order.
We begin with a sanity check of 10 distractors, involving three members from our team. The members reach a full consensus on the 10 samples. After reaching calibration, two team members proceeded to label 100 more distractors (50 each). The annotators found that 10 paragraphs could not have been made into good distractors (as they contain no dependencies). Out of the rest, 89\% of the generated distractors were correct. For the wrong ones, in 5 samples the generated paragraph was not \emph{coherent}, and in 5  the choice of events to replace was wrong. 

We deduce the distractor generation is effective, and create distractors for both gold and silver sets.

\section{Dataset Analysis}
\label{sec:datasets_analysis}

Our \emph{gold-set} contains 310 positives (analogies), each with one corresponding simple (random) distractor and one challenging distractor. Our \emph{silver-set} contains 403 positives, again with corresponding  distractors. 
We note this is a proof-of-concept, and it is possible to construct larger sets if desired. 


\noindent\textbf{Gold-set analysis.}
%
%
We first computed the distribution of close and far analogies (based on majority vote). When all three annotators voted positive, 40\% were far analogies. When at least one voted positive, the number increased to 47\%. We conclude that our dataset is relatively balanced between close/far analogies. Not surprisingly, disagreements are more common for far analogies.

Table~\ref{tab:rejection_distribution} shows different issues raised with the candidates. The most common is ``dissimilar relations'', indicating that GPT-3.5 has difficulties generating the relations. We note it is also quite easy  for annotators to detect. 
``Other'' was chosen in approximately a quarter of the cases. An example reason provided is \emph{inconsistent mapping} of entities. 
See Appendix~\ref{subsec:appendix_dataset_analysis} for more reasons.

\noindent\textbf{A Note on Scalability.}
The \emph{silver-set} is generated automatically at scale.
Our major annotation effort was to create the \emph{gold-set}.
For future users of the pipeline, we recommend the automatic route, with  short annotation rounds for quality assessment.

\noindent\textbf{A Note on Additional Data Released.}
\label{subsec:dataset_summary}
%
Through the different stages of the pipeline, we collect  information about candidates that does not make it into our gold or silver sets. We believe that this information might be beneficial for further research in this area. For example, \emph{differences} in judgments might be interesting. In addition, our human annotators give \emph{structured} feedback
(see Section \ref{subsec:humanInitialAnnotation}). If the annotators identified an issue with the generated relations, for example, it could still be the case that the \emph{paragraphs themselves} are analogous (which is the reason we do not use them as negatives). 
Thus, we decide to make this data available to the community.
We believe it opens up interesting avenues, from creating new types of distractors to teaching models how to automatically fix flawed analogies.


\section{Evaluating Humans and LLMs}

We use the data to develop the ProPara-Logy \emph{benchmark} of analogy recognition. 
We propose \emph{binary classification} and \emph{multiple-choice} tasks. We evaluate the performance of both {humans} and {state-of-the-art models}, experimenting in {zero-shot} and {guided} settings (using labeled examples). Our research questions are:

\label{sec:experiments}

\label{subsec:research_questions}

\noindent\textbf{RQ1:} How well can {humans} and {models} recognize analogies?

\noindent\textbf{RQ2:} Is the \emph{silver-set} useful for training models and improving their performance?

\noindent\textbf{RQ3:} Can the {distractors} fool humans and models?

\subsection{Tasks}
\label{subsec:tasks}
We propose two tasks. \emph{Binary classification} offers a simple and clean formulation;
  \emph{multiple-choice} is more similar to standardized test questions, adding an aspect of \emph{ranking} among choices.

\begin{table}[t!]
    \centering
    \begin{tabularx}{\columnwidth}{X *{5}{>{\centering\arraybackslash}X}}
        \toprule
        \textbf{Votes} & \textbf{Size} & \textbf{Drel} & \textbf{Minfo} & \textbf{Cyclic} & \textbf{Other} \\
        \midrule
        $=$ 0 & 443 & 93\% & 16\% & 21\% & 22\% \\
        $\leq$ 1 & 540 & 85\% & 23\% & 19\% & 27\% \\
        $\leq$ 2 & 648 & 73\% & 28\% & 17\% & 29\% \\
        \bottomrule
    \end{tabularx}
    \caption{Distribution of issues raised, by \#positive annotations:  dissimilar relations (Drel), misinformation (Minfo), cyclic vs. non-cyclic (Cyclic), and other. Annotators could choose more than one (hence sum $>$100\%). Most of the issues are with (LLM-generated) relations.}
\label{tab:rejection_distribution}
\end{table}

\begin{table*}[!t]
    \centering
    \begin{tabular}{ccccccc}
        \toprule
        \textbf{Row} & \textbf{Settings} & \textbf{Method} & \textbf{Overall} & \multicolumn{3}{c}{\textbf{Per Target Type}} \\
        \cmidrule(lr){5-7}
        &&&& \multicolumn{1}{c}{\textbf{Positives (50\%)}} & \multicolumn{2}{c}{\textbf{Negatives (50\%)}} \\
        \cmidrule(lr){5-5}
        \cmidrule(lr){6-7}
        &&&& \textbf{Analogy} & \textbf{Random} & \textbf{Distractor} \\
        \midrule
        1 & \multirow{7}{*}{Zero-shot} & Random Guess & 50 & \textcolor{graytext}{50} & \textcolor{graytext}{50} & \textcolor{graytext}{50} \\
        \hline
        2 && GPT4 & \textbf{79.5} & \textcolor{graytext}{95.2} & \textcolor{graytext}{92.9} & \textcolor{graytext}{34.8} \\
        3 && ChatGPT & 68.2 & \textcolor{graytext}{53.5} & \textcolor{graytext}{96.8} & \textcolor{graytext}{69.0} \\
        4 && Gemini Pro & 73.9 & \textcolor{graytext}{79.7} & \textcolor{graytext}{100} & \textcolor{graytext}{36.1} \\
        5 && FlanT5-XXL & 61.1 & \textcolor{graytext}{28.1} & \textcolor{graytext}{100} & \textcolor{graytext}{88.4} \\
        6 && FlanT5-XL & 59.7 & \textcolor{graytext}{25.1} & \textcolor{graytext}{100} & \textcolor{graytext}{88.4} \\
        7 && FlanT5-small & 49.3 & \textcolor{graytext}{0} & \textcolor{graytext}{97.4} & \textcolor{graytext}{100} \\
        8 && Humans & 79 & \textcolor{graytext}{58} & \textcolor{graytext}{100} & \textcolor{graytext}{100} \\
        \midrule
        9 & \multirow{3}{*}{Guided} & GPT4 (in-context) & 78 & \textcolor{graytext}{86.5} & \textcolor{graytext}{98.1} & \textcolor{graytext}{40.7} \\
        10 && FlanT5-small (fine-tune) & 74.4 & \textcolor{graytext}{87.1} & \textcolor{graytext}{96.1} & \textcolor{graytext}{27.1} \\
        11 && Humans & \textbf{92.5} & \textcolor{graytext}{95} & \textcolor{graytext}{100} & \textcolor{graytext}{80} \\
        \bottomrule
    \end{tabular}
     \caption{{The \textbf{Overall} and \textbf{Per Target Type}  \textbf{Accuracy (\%)}  of \emph{LLMs} and \emph{humans} in \emph{zero-shot} and  \emph{guided} settings, on the \emph{binary classification} task, evaluated on the \emph{gold-set}. Out of the models, GPT4 achieves the best overall accuracy \textbf{(row 2)}. Humans achieve better performance than models ($\sim$13\% gap in \textbf{Overall Accuracy}) after a cycle of learning from their mistakes \textbf{(row 2 vs. row 11)}. Interestingly, FlanT5 models tend to output ``not analogy'' more than GPT-4, rendering their performance on the true negatives higher, but overall their performance is worse \textbf{(rows 5-7 vs. row 2)}, see \textbf{Section~\ref{subsec:results}, RQ1}. The training of FlanT5-small on the \emph{silver-set} significantly improved its \textbf{Overall Accuracy} \textbf{(row 10 vs. row 7)}, 
    see \textbf{Section~\ref{subsec:results}, RQ2}.
    Comparing challenging negatives (\textbf{Distractor}) with simple negatives (\textbf{Random}), we observe a performance decline in both humans and LLMs \textbf{(rows 2-6, 8-11)}, except for FlanT5-small, which almost always predicts ``not analogy''. This reduction is statistically significant for models but not for humans.}}
    \label{tab:binary_classification_results}
\end{table*}

\begin{table}[!ht]
    \centering
    \footnotesize
   \begin{tabular}{ccccc}
        \toprule
        \textbf{Row} & \textbf{Settings} & \textbf{Method} & \textbf{Basic} & \textbf{Advanced} \\
        \midrule
        1 &  & Random Guess & 25 & 25 \\
        \hline
        2 & \multirow{6}{*}{Zero-shot} & GPT4 & \textbf{95.5} & \textbf{83.2} \\
        3 && ChatGPT & 74.2 & 59 \\
        4 && Gemini Pro & 87.4 & 62.6 \\
        5 && FlanT5-XXL & 87.4 & 75.2 \\
        6 && FlanT5-XL & 68.4 & 55.5 \\
        7 && FlanT5-small & 32.9 & 32.9 \\
        \midrule
        8 & Guided & Humans  & \textbf{100} & \textbf{96} \\
        \bottomrule
    \end{tabular}
    \caption{{The \textbf{Accuracy (\%)} of \emph{LLMs} and \emph{humans} in \emph{zero-shot} and \emph{guided} settings, on the \emph{multiple choice} task, evaluated on the \emph{gold-set}. The \textbf{Basic} setting uses simple negatives (random), while the \textbf{Advanced} includes challenging negatives (distractors). Humans achieve better performance than models ($\sim$13\% gap in the advanced setup); out of the models, GPT4 achieves the best results \textbf{(row 2)}, see \textbf{Section~\ref{subsec:results}, RQ1}. Distractors reduce performance in both humans and LLMs. This decline is statistically significant for the models, but not for humans \textbf{(rows 2-8 Advanced vs. Basic)}, see \textbf{Section~\ref{subsec:results}, RQ3}. We additionally confirmed that the proportion of mistakes due to choosing the challenging distractor is much higher. For more details, see \textbf{Section~\ref{subsec:results}, RQ3}}. Note that here we show the results of models only in zero-shot, as we already addressed \textbf{RQ2} -- ``Is the silver-set useful for training models?'' in Table~\ref{tab:binary_classification_results}. We leave training of models for the \emph{multiple choice} task for future work.}
    \label{tab:multiple_choice_results}
\end{table}

\noindent\textbf{Binary classification.}
Given a pair of paragraphs base \base and target \targ, each describing a scientific process in natural language, the task is to decide whether the processes are analogous. The target paragraph could either be: 
(1) \textbf{Analogy} (positives), (2) \textbf{Random} ProPara paragraphs with a different subject than \base (simple negatives, see footnote \ref{random_paragraph_footnote})
 or (3) \textbf{Distractor} paragraphs  (challenging negatives, see Section \ref{subsec:distractors}).
In the benchmark, we balance the samples s.t. 50\% of target paragraphs are analogies, 25\% are simple negatives and 25\% are distractors. 

\noindent\textbf{Multiple-choice.}
Given a base paragraph \base, along with four candidate paragraphs, the task is to identify the paragraph that is most analogous to \base. We use two different setups. (1) \textbf{Basic}: candidates are one analogous paragraph and 3 random paragraphs. (2) \textbf{Advanced}: In this setup, we increase the difficulty by including the distractor corresponding to the correct answer. 
However, this results in always having two extremely similar candidates (the analogous paragraph and its distractor), and both trained models and humans might realize that the correct answer always lies between them. 
To overcome this issue, we generate distractors both for the correct answer and for the random paragraph, and use them as our four candidates. This way, candidates include \emph{two} pairs of similar paragraphs.

\subsection{Baselines}
\label{subsec:methods}

We evaluate both \emph{state-of-the-art LLMs} and \emph{humans} in {zero-shot} and {guided} settings.

\noindent\textbf{Models.}
We tested ChatGPT\footnote{https://chat.openai.com/chat}, GPT-4 \cite{openai2023gpt}, Gemini Pro \cite{team2023gemini},
FlanT5-small, FlanT5-XL, and FlanT5-XXL \cite{chung2022scaling}, all with their official implementations and default parameters. See Appendix~\ref{subsec:appendix_experiments} for additional models, which performed poorly.

{The families of models we have experimented with represent state-of-the-art.
Our task poses significant NLP challenges, and recent work by \citet{sultan-shahaf-2022-life} suggests that more traditional models such as SBERT \cite{reimers-2019-sentence-bert} could only identify very close analogies with similar entities. Thus, we decided to focus on recent state-of-the-art LLMs and leave testing of additional models for future work.}


We start experimenting in a \emph{zero-shot} setting\footnote{We note that while GPT4 is used in the pipeline {(\S\ref{subsec:analogyFiltering})}, 
its parameters have not been updated in the process. See https://platform.openai.com/docs/models/gpt-3}.
In the binary task, we use 620 instances (310 analogies, 155 distractors and 155 random) from our \emph{gold-set}. In the multiple-choice we use the 310 analogous paragraphs as one of the candidates, adding three random paragraphs (basic setup), or a distractor, a random paragraph and a distractor generated for it (advanced setup). 
See prompts in Appendix~\ref{subsec:appendix_experiments}.

In addition to the zero-shot setting, we experimented with a \textbf{guided setting} to improve the performance of models and humans in the binary task using labeled examples.
Where we could fine-tune models, we did. Other times, such as with GPT4 (the best model from zero-shot), we used few-shot examples. We experimented with several prompts, based on successes and failures of the model. Overall accuracy remained similar, and we chose a prompt that includes five mistakes (3 distractors, 1 analogy, 1 random); the rationale was to include more examples of common mistakes. See Appendix~\ref{subsec:appendix_experiments}.

\noindent\textbf{Humans.}
In addition to the evaluation of LLMs, we are also interested in assessing the performance of humans on both tasks. We employ new AMT workers, who had not participated in creating the dataset. In both tasks, every instance is evaluated by 3 annotators.
We publish the majority vote accuracy, and agreement as the \% of perfect agreement.
See Appendix~\ref{subsec:appendix_experiments} for task instructions.

On the binary task, we run the experiment in two stages, mimicking the zero-shot and guided settings of the models. In the zero-shot setting, we show the crowdworkers 100 randomly sampled instances from the \emph{gold-set}, including 50 positives (equally divided into close and far analogies), 25 simple negatives and 25 challenging distractors. 

For the guided setting, we show workers examples based on their errors (similar to what we did with GPT-4). Then, we use another set of samples (with different base paragraphs) with 10 {close analogies}, 10 {far analogies}, 10 {simple negatives}, and 10 {distractors}.
For the multiple-choice task, we show 25 instances for the basic setup, and {another} 50 for the advanced setup (using different base paragraphs).
See Appendix~\ref{subsec:appendix_experiments} for the tasks.

\subsection{Results}
\label{subsec:results}

{Our results are summarized in Table~\ref{tab:binary_classification_results} for the \emph{binary classification} task, and Table~\ref{tab:multiple_choice_results} for the \emph{multiple choice} task.}

\noindent\textbf{RQ1: What is the performance of humans and models?}
In the binary task in zero-shot, GPT4 achieves the highest overall accuracy of 79.5\%, succeeding on analogies and simple negatives but struggling with distractors. 
Gemini Pro follows with overall accuracy of 73.9\%, then
ChatGPT with overall accuracy of 68.2\%. Not surprisingly, we can also see that Flan models get better as they grow bigger.

Humans achieve 79\% overall accuracy ($\sigma = 0.04$), nearly matching the best model. Interestingly, humans achieved perfect accuracy on simple negatives and distractors, but were too strict and ruled out many correct analogies. Agreement was 70\% (random chance 25\%). 
Initially, we expected humans to outperform models. Thus, we set out to explore whether adding a guidance step helps.
For the guided settings, we used the best model (GPT4) with few-shot examples of its mistakes. Similarly, we showed the crowdworkers their mistakes. We found that humans were able to improve significantly, achieving an overall accuracy of 92.5\% ($\sigma=0.014$) and agreement of 80\%.
{We conclude the task is complex, but possible to explain.}
On the other hand, GPT4's performance is similar, even testing numerous prompt variations (Section \ref{subsec:methods}). 

We note that this task is harder than the annotation task of Section \ref{subsec:humanInitialAnnotation}. Here, annotators only see the paragraphs (not the similar relations, subjects, or domains). Additionally, they have to decide whether the paragraphs are analogous, as opposed to going over a structured list of potential issues.

In the \emph{multiple-choice} task, the best model is again GPT4, achieving overall accuracy of 95.5\% (basic setup) and 83.2\% (advanced setup). 

For the multiple-choice task, we employed the same annotators from the binary task after guidance.
In the basic setup, humans (majority vote) achieve a perfect accuracy of 100\% ($\sigma=$0.04), and agreement of 88\%.
In the advanced setup, an accuracy of 96\% ($\sigma=0.04$), and agreement of 66\% (chance agreement is 6.25\%). 

To conclude, humans achieve better performance than models ($\sim$13\% gap) after light supervision; out of the models, GPT4 achieves the best results.

\noindent\textbf{RQ2: Is the \emph{silver-set} useful for training models?}
We employ FlanT5-small, which is a small model of only 80M parameters, fine-tune it on the \emph{silver-set} (which was automatically generated) for the binary classification task, and test it on the \emph{gold-set}.  We choose FlanT5-small to test whether high accuracy can be achieved even with a small model.
We use the same prompt from the zero-shot setting. See Appendix~\ref{subsec:appendix_experiments} for details about training.

FlanT5-small's overall accuracy improved from 49.3\% to 74.4\% after fine-tuning, surpassing even the largest Flan model (FlanT5-XXL), in zero-shot (see Table \ref{tab:binary_classification_results}). This result is statistically significant with a p-value of 1.3e-06 in the McNemar test, at the 0.05 level with Bonferroni correction.


\noindent\textbf{RQ3: Are the distractors effective?}
In the binary classification task, we can see that both humans and LLMs (except FlanT5-small, which almost always predicted ``not analogy'') achieve nearly perfect accuracy on the simple negatives, but lower accuracy on the challenging distractors (see Table \ref{tab:binary_classification_results}).
In the multiple-choice task, we can see a drop in performance for both LLMs (except FlanT5-small) and humans when transitioning from basic setup without distractors to advanced with distractors (see Table \ref{tab:multiple_choice_results}).
We use the McNemar test to assess statistical significance, reaching p-values of 7e-08 for GPT4, 4.3e-14 for Gemini Pro, 6.3e-06 for ChatGPT, 1.5e-05 for FlanT5-XXL, and 0.0009 for FlanT5-XL (all statistically significant at the 0.05 level after Bonferroni correction). The drop in accuracy for humans was not significant.

Next, we compute the \emph{percentage} of errors resulting from incorrectly choosing the distractor: In humans it is 100\%, for
GPT4 92.3\%, Gemini Pro 75.0\%, ChatGPT 66.9\%, FlanT5-XXL 62.3\%, FlanT5-XL 25.4\%, and FlanT5-small 40.5\% (random chance 25\%). Thus, LLM mistakes mainly stem from selecting the distractor. In the case of humans, the \emph{absolute} number of mistakes is quite small, so we cannot draw a firm conclusion.



\section{Conclusions}

Analogy-making is crucial for AI to generalize and adapt to unfamiliar contexts. 
%
We designed a pipeline, \emph{ParallelPARC}, leveraging  LLMs to generate complex analogies and distractors. 
We demonstrated our pipeline by creating \emph{ProPara-Logy}, a dataset of analogies between scientific processes.
\emph{ProPara-Logy} is orders of magnitude larger than previous datasets of full paragraphs, and could easily be expanded via the pipeline.

Our experiments show humans outperform models after light supervision, and that even the best models are more sensitive to distractors than humans.
We also show that  automatically generated data is useful for training and improving models.

{Our pipeline is easy to adapt to new domains, requiring only small changes in the prompts. We hope researchers will use it in domains where analogies have shown promise. For example, in education \cite{Duit1991OnTR,  Clement1993UsingBA}, analogies can be used to leverage students' existing knowledge to make abstract or challenging material easier to grasp (e.g., in biology, the heart is frequently compared to a pump to help students understand how it circulates blood throughout the body, similar to how a pump moves water);  in computer-assisted creativity \cite{Moreno2014FundamentalSI, Hope2017AcceleratingIT}, analogies can be used to inspire designers and engineers in solving new problems by using existing ideas from another field (e.g., NASA has embraced the principles of origami to develop foldable solar panels and satellite antennas).}
We hope this work will spur more NLP work on analogies, leading to novel tasks and benchmarks.

\paragraph{Acknowledgements}
We would like to thank the HyadataLab, NLP-HUJI, BIU-NLP and SheffieldNLP for their valuable feedback.
We also thank the anonymous reviewers for their constructive comments. 
This work was supported by the European Research Council (ERC) under the European Union’s Horizon 2020 research and innovation programme 
(grant no. 852686, SIAM).

\section*{Ethical Considerations}

\noindent\textbf{Misuse of analogies.}
Research has revealed that people often find it difficult to discern nuances or limitations in presented analogies \cite{holyoak1996mental}.
For example, in \citet{Swain2000TheWA} an analogy is used to explain medical students the intricacies of the cardiovascular system by likening it to a city water supply.
However, this analogy might also confuse them, as it fails to acknowledge crucial distinctions between water and blood, such as the existence of blood clots.
Thus, one might wish to alert people who read analogies generated by our pipeline to this possibility, as well as the possibility of LLM hallucinations.

\noindent\textbf{Crowdsourcing.}
Human annotations and evaluations were carried out through crowdsourcing (Amazon Mechanical Turk platform).
The workers are native English speakers from the US.
Workers were compensated at a rate of \$15 per hour (higher than the minimum wage in their states). We set the price per HIT by calculating the average completion time for sample HITs.

\noindent\textbf{Dataset.}
We used the ProPara dataset of paragraphs describing scientific processes in English, taking 390 titles from its training set (allenai.org/data/propara)\footnote{https://github.com/allenai/propara (Apache-2.0 license, no explicit intended use)}
and generated the ProPara-Logy dataset. 
We removed all content in the ProPara-Logy that might contain  information about the annotators, such as worker IDs.
Note that our generated dataset focus is on the scientific domain, limiting cultural or situational biases.

\noindent\textbf{Computation.} 
Zero-shot experiments require about an hour to run both tasks on an NVIDIA A100 GPU, with the majority of the time spent on interactions with the GPT model's API. These experiments are conducted using Google Colab Pro+ on the Ubuntu version of Linux. Fine-tuning experiments, involving both training and inference of FlanT5-small, take less than 15 minutes on an NVIDIA RTX 6000 GPU. These experiments are run from the university cluster, operating on Debian GNU/Linux.

\section*{Limitations}




\noindent\textbf{Relying on closed models (e.g., OpenAI models).} In closed models, the architecture, training data, and training methodologies are not available; furthermore,  these models belong to a company and thus might be shut down or deprecated in the future.
Nevertheless, these models are considered to be state-of-the-art, are widely in use and have gained significant attention from both experts and non-experts. Thus, we believe it is valuable to use them in this work, acknowledging their limitations.

\noindent\textbf{Sensitivity to prompts.} It is known that LLMs are sometimes sensitive to small changes to the prompts.

\noindent\textbf{Domains.} In this work we focused on generated data for scientific processes across several (specific) domains. The results in other domains are yet to be explored.

\noindent\textbf{Language.} Our benchmark contains solely English texts. The results may differ in other
languages.

\bibliography{anthology,custom}

\appendix

\section{Appendix}

\subsection{Reproducibility}

\subsubsection{Models}
The models we used for evaluation are detailed in Section~\ref{subsec:methods}. 
Regarding the zero-shot experiments: after loading the models, it takes approximately one hour to run the models on both tasks on an NVIDIA A100 GPU. 
The majority of this duration is attributed to interactions with the GPT model's API.
We run it using Google Colab Pro+ (the operating system is the Ubuntu version of Linux).

Regarding the fine-tuning experiments: both training and inference of FlanT5-small took less than 15 minutes on NVIDIA RTX 6000 GPU. We run it from the cluster of the university  (the operating system is: Debian GNU/Linux).
Trained models hyper-parameters (and the range of values we tried) are provided in Appendix~\ref{subsubsec:appendix_flanT5_fine_tune}. Full implementation is provided in the attached code.
\label{subsec:appendix_models_running_details}

\subsubsection{Statistics}
The details about the pipeline generation are provided in Section~\ref{sec:dataset}. Dataset Statistics are provided in Section~\ref{sec:datasets_analysis}. 
A link to a downloadable version of the dataset is available in the code. 
A complete description of the annotation process is provided in Sections~\ref{subsec:humanInitialAnnotation}, and \ref{subsec:humanAnnotation}. 

\subsubsection{Code}
The attached code includes the full implementation, dependencies, training code, evaluation code, pre-trained models, README files, and commands necessary to reproduce the results presented in the paper.

\subsection{Analogy Candidates Generation}
\label{subsec:appendix_analogy_candidates_generation}
See Figures~\ref{tab:analogy_candidates_generation_prompt1}, and \ref{tab:analogy_candidates_generation_prompt2} for our solution using the two prompts.
See Figure~\ref{fig:gpt3_one_prompt_problems} for an example of what happens when we used one prompt for the whole task.
\subsubsection{Model's parameters}
For generating the analogy candidates, we use GPT-3.5 (text-davinci-003) \cite{Brown2020LanguageMA} with temperature=0.7, max\_tokens=1000, and top\_p=1.
\label{subsubsec:appendix_analogy_candidate_generation_model_params}

\subsection{Automatic Filtering and Labeling}
\label{subsec:appendix_filtering}
See Figure~\ref{tab:auto_label_prompt} for the prompt given to the auto-labeling model.
\subsubsection{Model's parameters}
For our auto-labeling model, we used GPT-4 \cite{openai2023gpt} with the following parameters: temperature=0.5, max\_tokens=4000, top\_p=0.
\label{subsubsec:auto_label_model_params}

\subsection{Human Annotation}
\label{subsec:appendix_human_annotation}
In this section we will give more details about the reasons for further inspection, and the annotation process.
\subsubsection{Reasons for further inspection}
\label{subsubsec:appendix_annoatation_further_inspection_reasons}
Here is the list of some popular reasons we found:
\begin{itemize}
\item \textbf{Dissimilar relations} -- when at least one line of relations consists of dissimilar relations. For example:
(precipitation, \emph{falls}, on the ground) like (rotor, \emph{rotates}, generator) contains a pair of relations with dissimilar meaning for the verbs ``falls'' and ``rotates''. 
\item \textbf{Misinformation} -- when one of the paragraphs (or the relations) contain misinformation. For example, one paragraph mentions ``rain droplets rise to the atmosphere'' instead of ``falls to the ground''. 
\item \textbf{Cyclic vs. non-cyclic process} -- when one paragraph describes a cyclic process and the other not (e.g, one paragraph about the water cycle process which is cyclic, and another on human digestive system, which is not cyclic). 
\item \textbf{Other} -- any other reason. 
\end{itemize}

\subsubsection{The annotation process}
\label{subsubsec:appendix_annoatation_process}
\begin{figure*}[t]
\centering
\includegraphics[width=.99\textwidth]{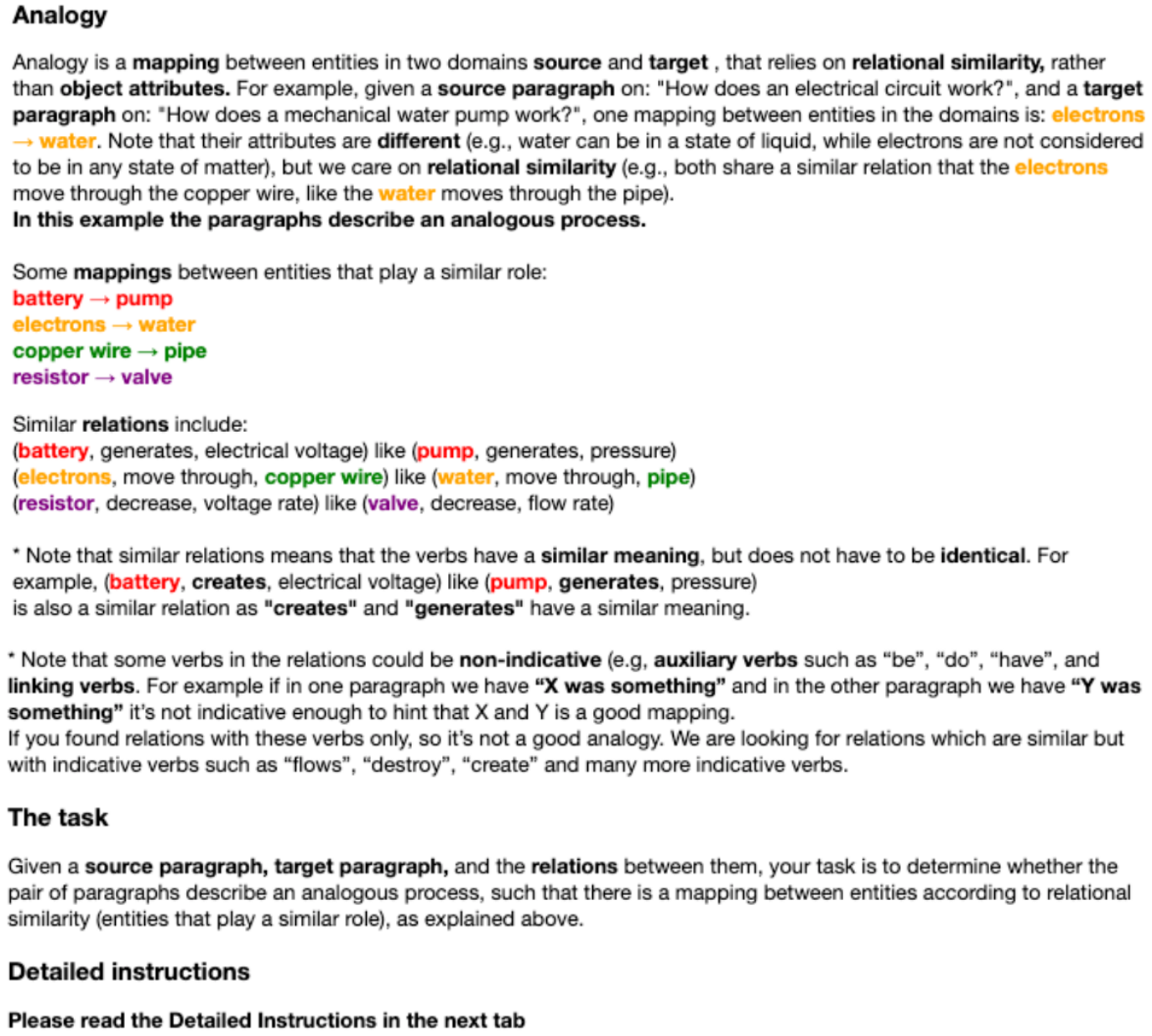}
\caption{The instructions screen of the annotators in Amazon Mechanical Turk. It includes a background on analogies, and explanation about the task.}
\label{fig:mturk_instructions_screen}
\end{figure*}

\begin{figure*}[t]
\centering
\includegraphics[width=.99\textwidth]{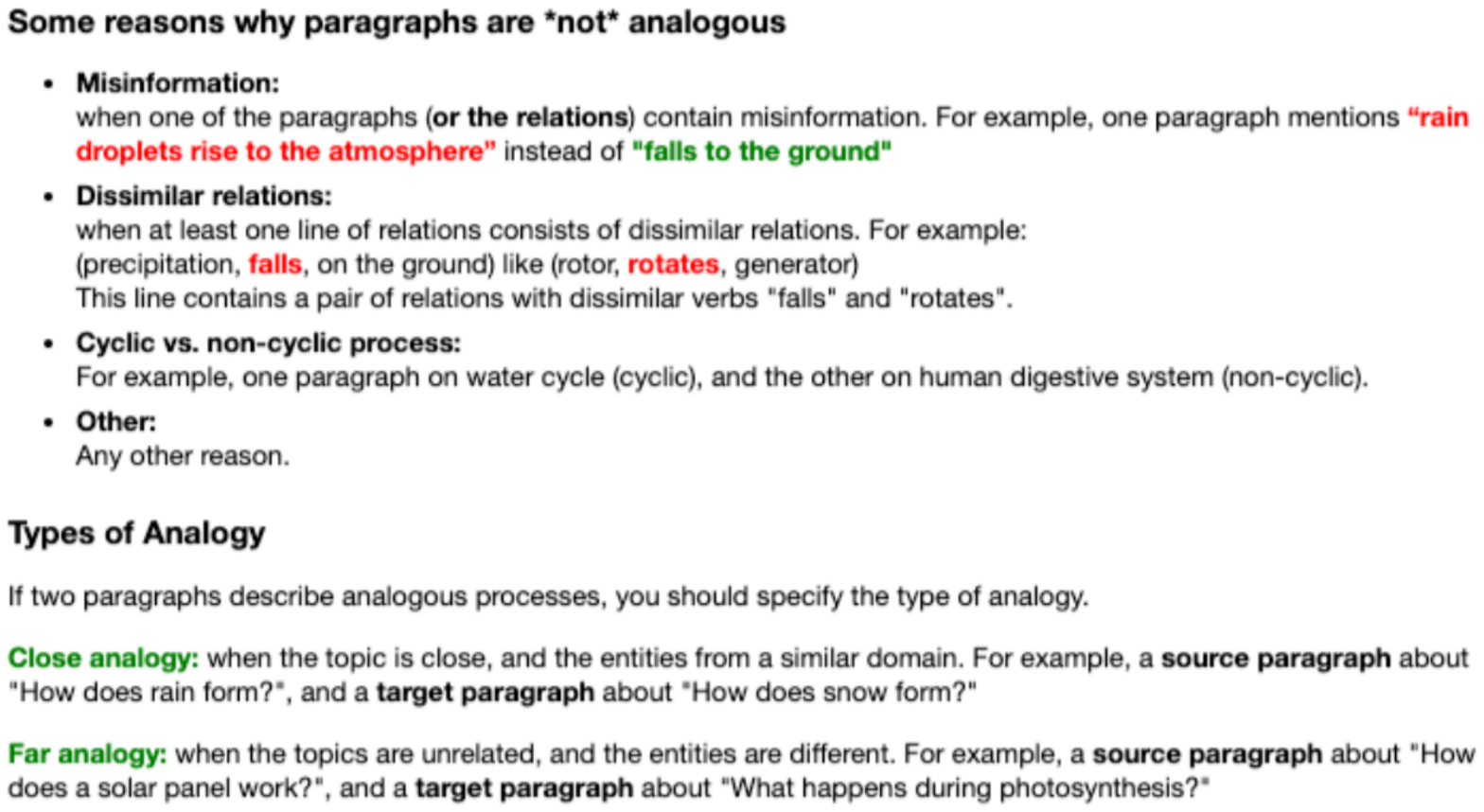}
\caption{The detailed instructions screen of the annotators in Amazon Mechanical Turk. It includes the reasons why a sample is currently not analogous (in its current form), hence is given for further inspection, as well as the types of analogy.}
\label{fig:mturk_detailed_instructions_screen}
\end{figure*}

\begin{figure*}[t]
\centering
\includegraphics[width=.99\textwidth]{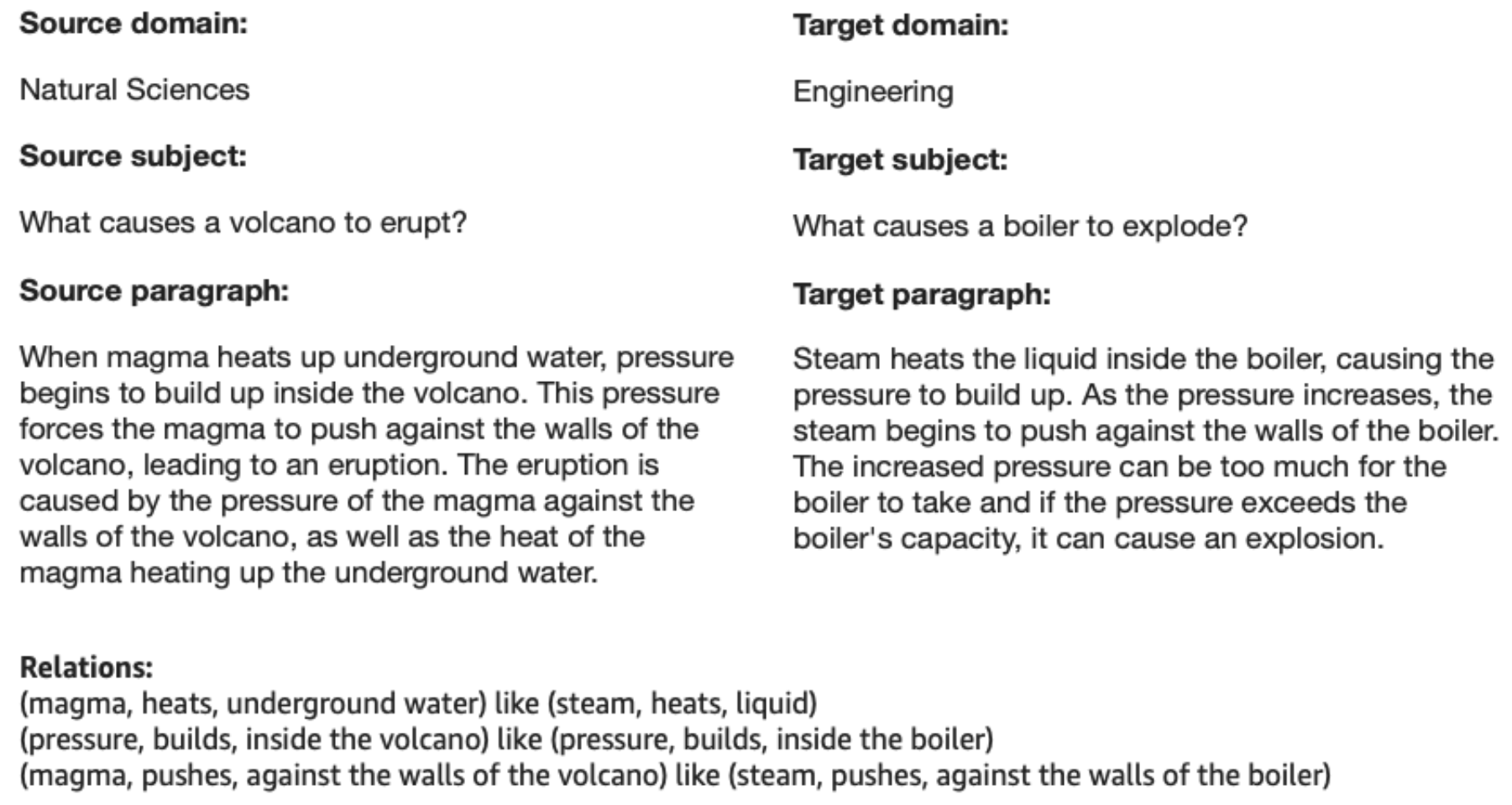}
\caption{An example for a sample from the \emph{analogy candidates}  given to the workers in the phase of instructions. This example is labeled as a \emph{far analogy}.}
\label{fig:mturk_instruction_example1}
\end{figure*}

\begin{figure*}[t]
\centering
\includegraphics[width=.99\textwidth]{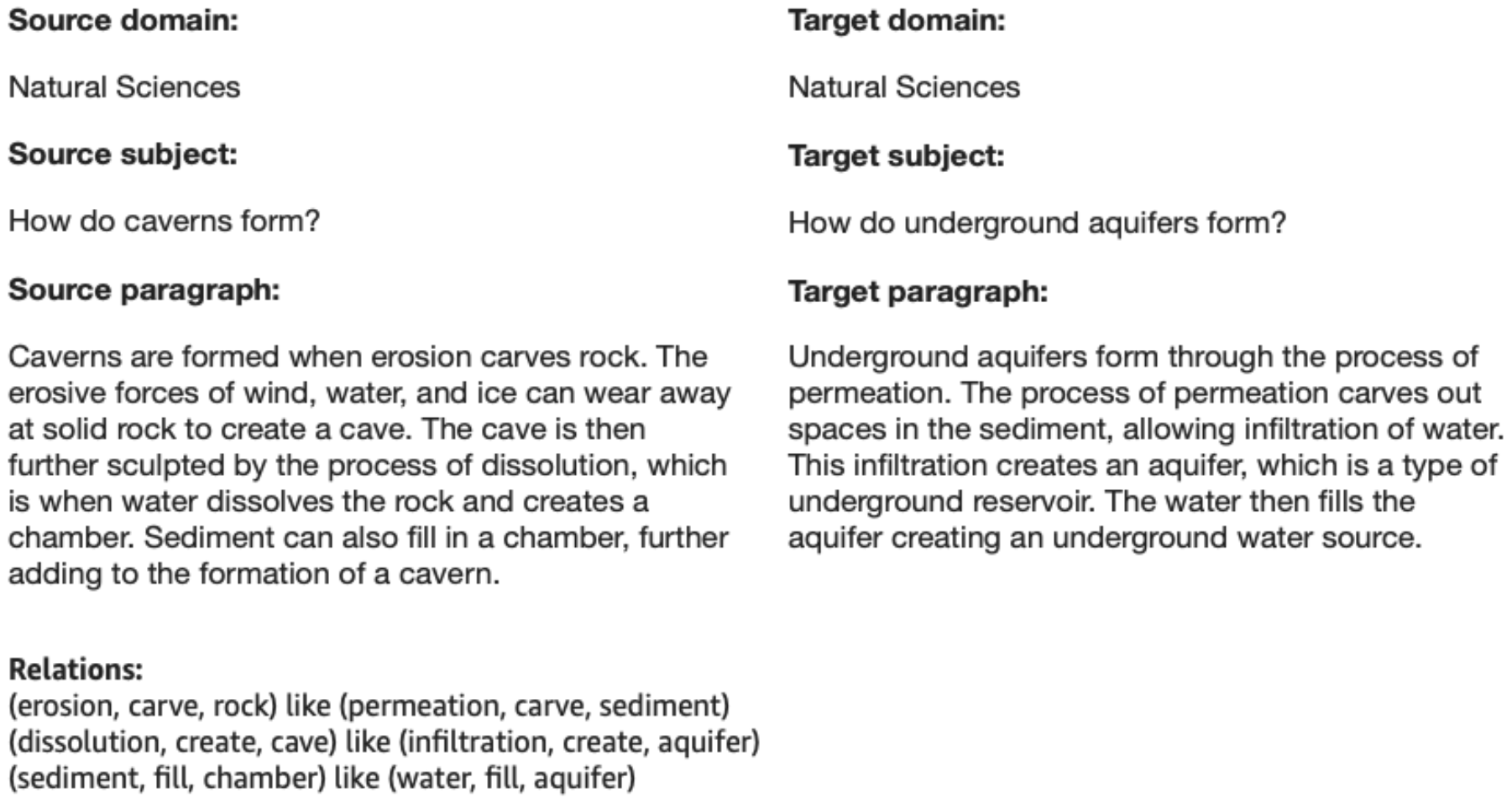}
\caption{An example for a sample from the \emph{analogy candidates}  given to the workers in the phase of instructions. This example is labeled as a \emph{close analogy}.}
\label{fig:mturk_instruction_example2}
\end{figure*}

\begin{figure*}[t]
\centering
\includegraphics[width=.99\textwidth]{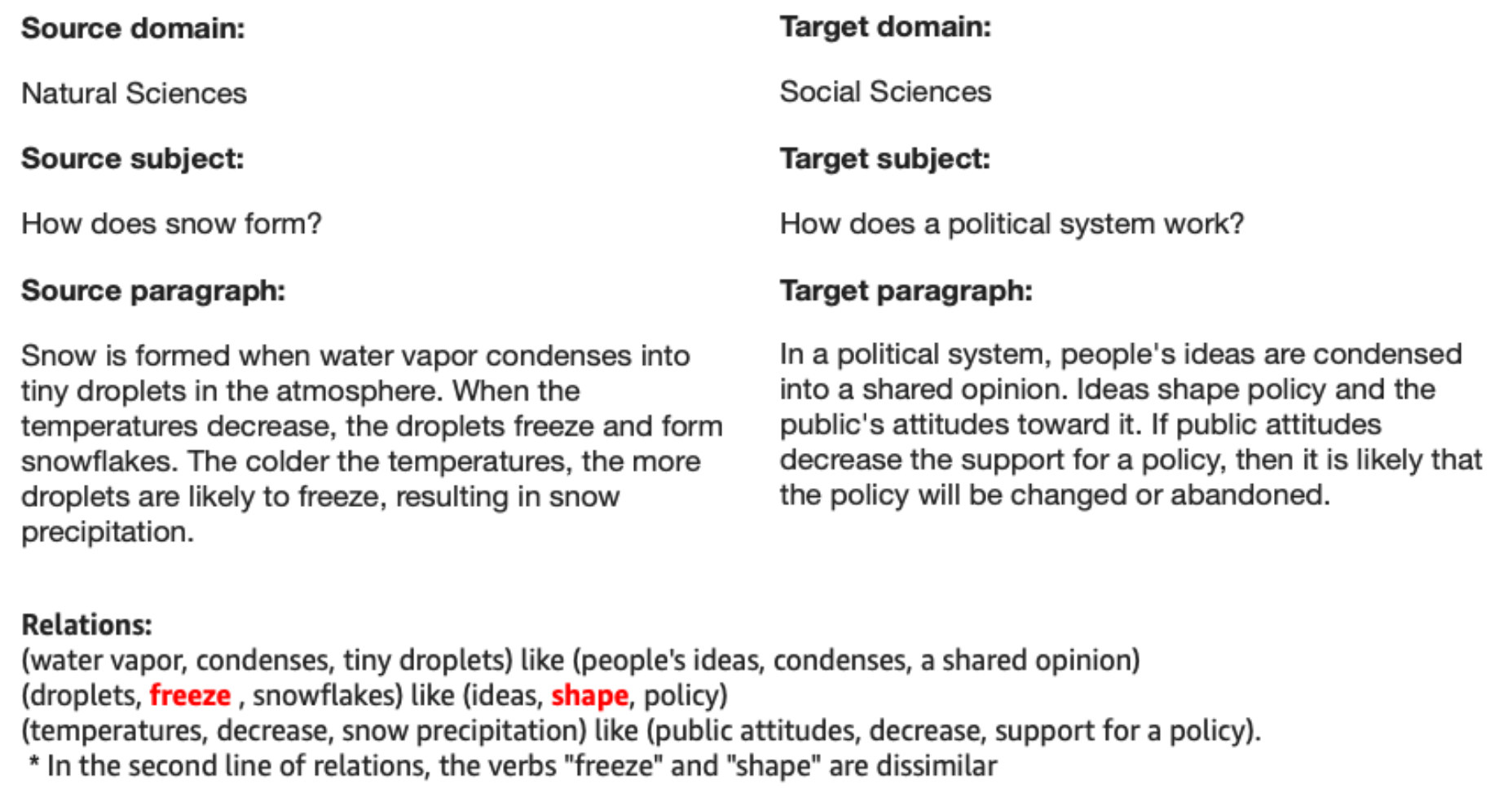}
\caption{An example for a sample from the \emph{analogy candidates} given to the workers in the phase of instructions. This example  is labeled with the reason of \emph{dissimilar relations}.}
\label{fig:mturk_instruction_example3}
\end{figure*}

\begin{figure*}[t]
\centering
\includegraphics[width=.99\textwidth]{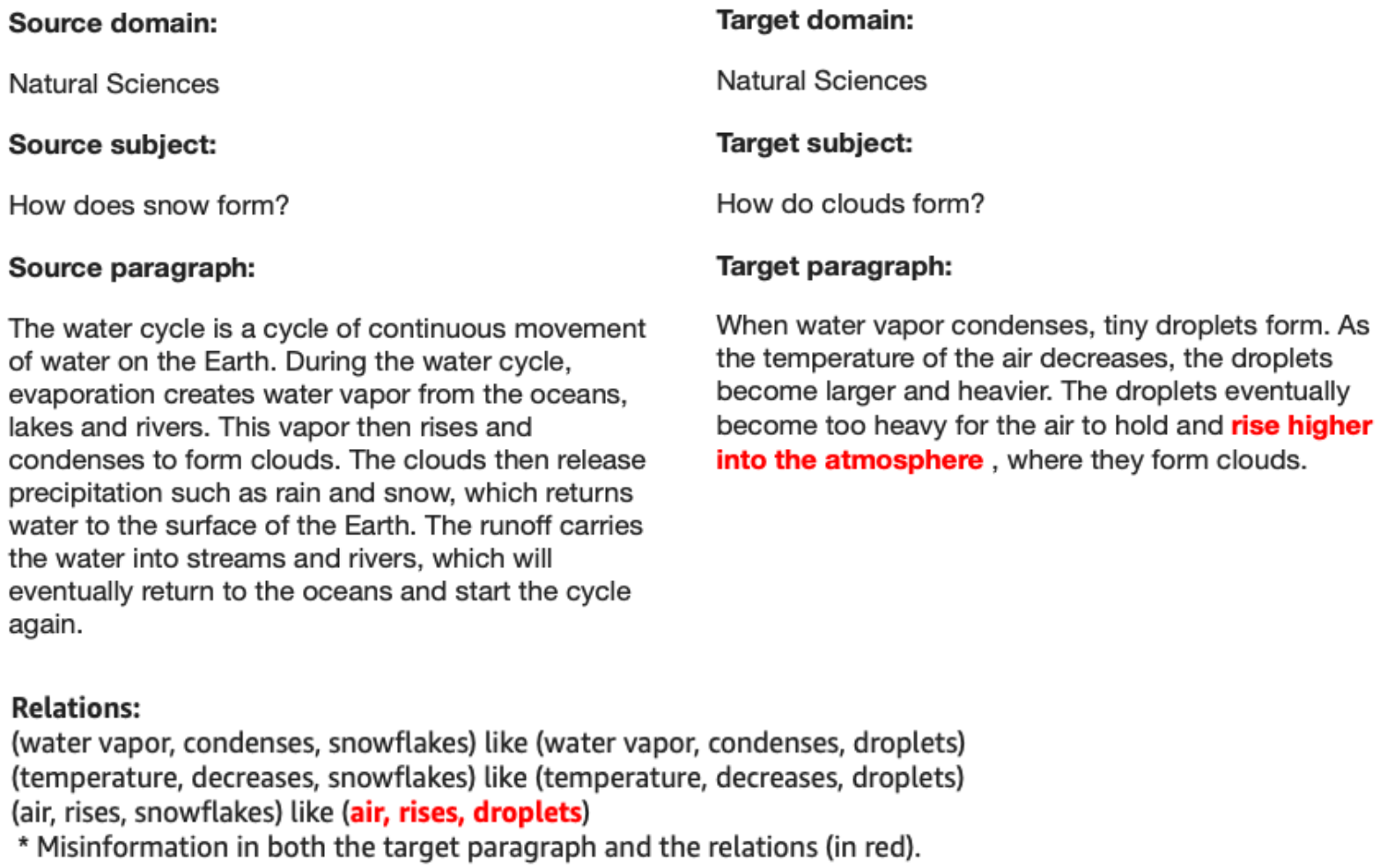}
\caption{An example for a sample from the \emph{analogy candidates} given to the workers in the phase of instructions. This example  is labeled with the reason of \emph{misinformation}.}
\label{fig:mturk_instruction_example4}
\end{figure*}

\begin{figure*}[t]
\centering
\includegraphics[width=.99\textwidth]{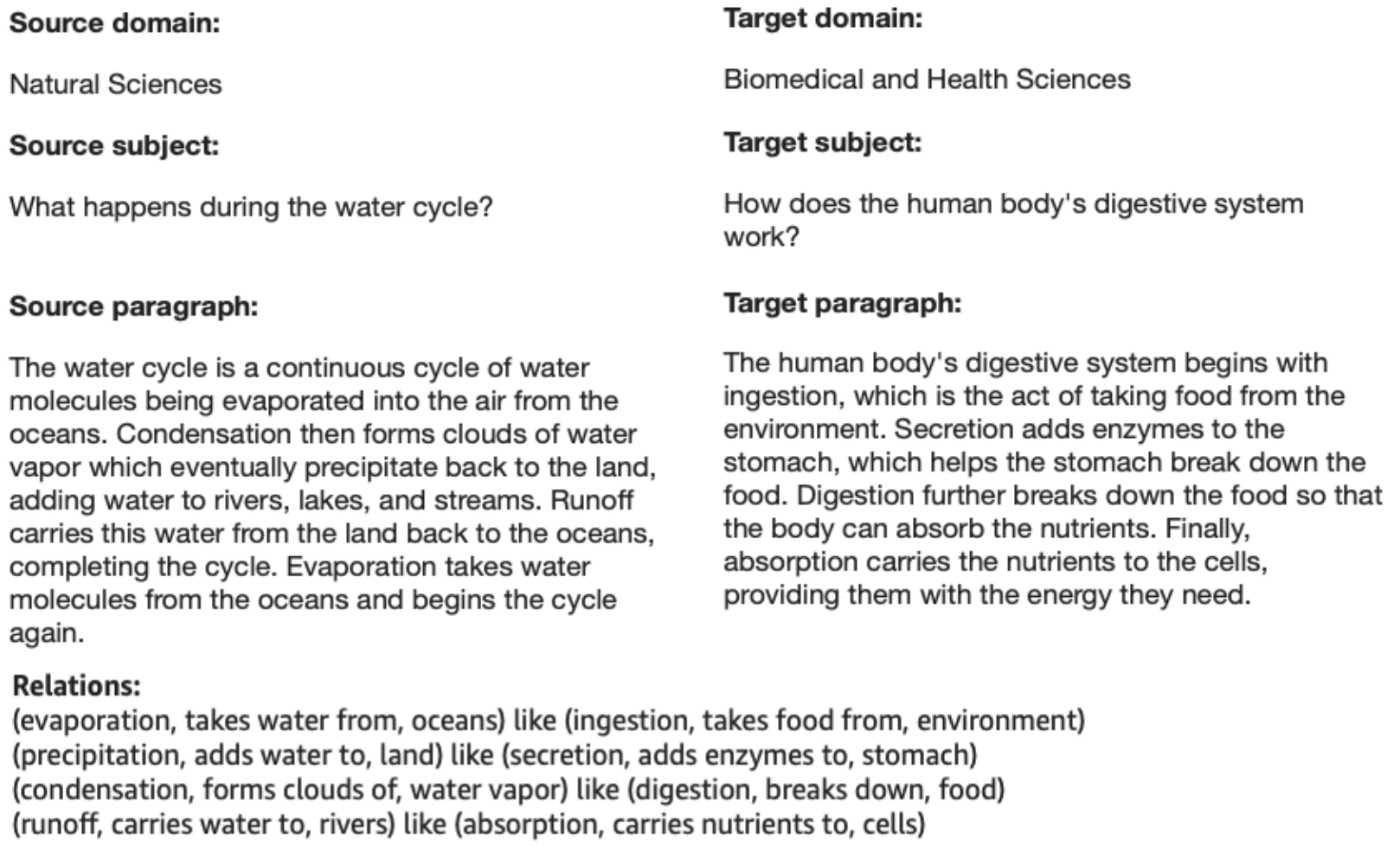}
\caption{An example for a sample from the \emph{analogy candidates} given to the workers in the phase of instructions. This example  is labeled with the reason of \emph{cyclic vs. non-cyclic process}.}
\label{fig:mturk_instruction_example5}
\end{figure*}

\paragraph{Human annotation task}
We start by giving the workers the \textbf{instructions} for the task, which include a background on analogies, explanation about the task and the labels. See Figure~\ref{fig:mturk_instructions_screen}, and Figure~\ref{fig:mturk_detailed_instructions_screen} for the instruction screens given to the workers in the Amazon Mechanical Turk platform. In addition to the instructions, we supplied 5 full examples (close analogy, far analogy, and 3 candidates for further inspection with different reasons). See Figures~\ref{fig:mturk_instruction_example1}, \ref{fig:mturk_instruction_example2}, \ref{fig:mturk_instruction_example3}, \ref{fig:mturk_instruction_example4} and \ref{fig:mturk_instruction_example5} for the five examples. 
After the workers read the instructions for the task, they performed a \textbf{qualification exam} consists of 10 samples (equally divided between analogies and rejected samples). 7 out of 12 workers passed our performance bar -- at least 8 out of 10 correct answers.
Then, the workers start to annotate \emph{analogy candidates}. The first phase is \textbf{initial annotation}, where our 7 highly-qualified workers labeled 130 samples from the \emph{analogy candidates}. We chose 30 random samples with their label of how many workers vote for analogy (between 0 and 3) to feed as in-context few-shot samples to the GPT-4 auto-labeling model.
\paragraph{Human validation}
The next phase is the \textbf{validation}, in which we run our GPT-4 auto-labeling model in batches from the \emph{analogy candidates}, and give the highly-qualified workers to label only samples where the model predicts full agreement. In this way, we filter the most probable analogies and candidates for further inspection.
\paragraph{Workers consent}
We obtained worker consent for all workers participating in the task. Workers have been told about the objective of the work, and how their annotations will be used. They have also been told they were annotating data generated by AI. 
Data collection has been approved by the Hebrew University board of ethics.


\subsection{Distractors Generation}
\label{subsec:appendix_distractors}
See Figures~\ref{tab:distractor_replace_events_one_shot} and \ref{tab:distractor_write_paragraph_few_shot} for the two prompts to generate distractors.
\subsubsection{Distractors formulation}
\label{subsubsec:appendix_distractors_formulation}
Here is the formulation of our distractors.
Let $x$ and $y$ be two events in \targ which describes an analogous process to \base, which is a paragraph from the ProPara dataset, in the form of procedural text. An \emph{event} in paragraph is usually described in 1–2 sentences. Let $t_{x}$ and $t_{y}$ be the timestamps of events $x$ and $y$ in \targ, such that $t_{x}<t_{y}$, and $x$ must happen before $y$, in other words $x$ is a prerequisite of $y$, or $y$ is dependent on $x$ which has to be presented before $y$. Our aim is to create a \emph{coherent} paragraph \targprime such that $y$ will be presented before $x$ in the sequence of events. This distractor paragraph will include similar \emph{first-order} relations, but dissimilar \emph{higher-order} relations, which result in different
\emph{cause-and-effect-relationships} and possibly make \targprime \emph{illogical}. 

\subsubsection{Model's parameters}
We use GPT-4 with temperature=1.0 (for the one-shot prompt of creating new events order) and temperature=0.00001 (for the second few-shot prompt of creating the distractor paragraph). We used the other default parameters for both prompts.
\label{subsubsec:generate_distractors_model_params}

\subsection{Dataset Analysis}
\label{subsec:appendix_dataset_analysis}
Here are the popular issues that annotators found as ``Other''.
\begin{itemize}
    \item \textbf{Inconsistent mapping}: when the mapping that can be inferred by the supplied relations is inconsistent, which means one entity in the base is mapped to more than one entity in the target
    \item \textbf{Incorrect structure of relation}: the correct format for relations is: (entity1, verb, entity2), but some generated candidates had a wrong format (e.g, (verb, verb, entity)).
    \item \textbf{Relations and paragraphs misalignment}
\end{itemize}

\subsection{Evaluating Humans and LLMs}
\label{subsec:appendix_experiments}
\begin{figure*}[t]
\centering
\includegraphics[width=.99\textwidth]{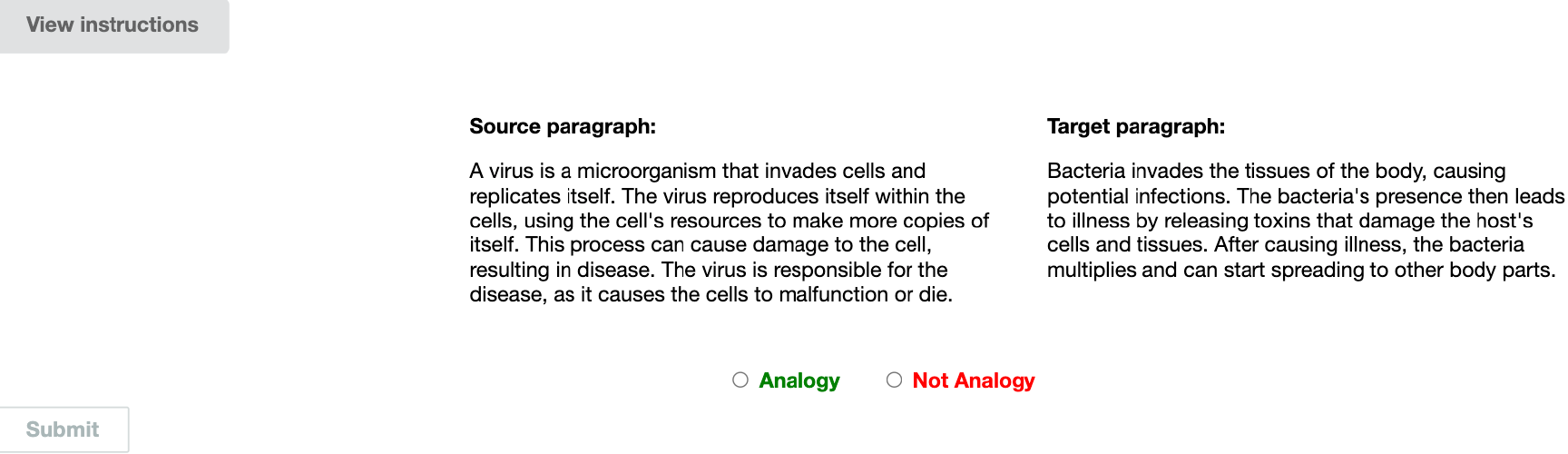}
\caption{The screen for the crowdworkers in AMT for the binary classification task.}
\label{fig:mturk_binary_classification_screen}
\end{figure*}

\begin{figure*}[t]
\centering
\includegraphics[width=.99\textwidth]{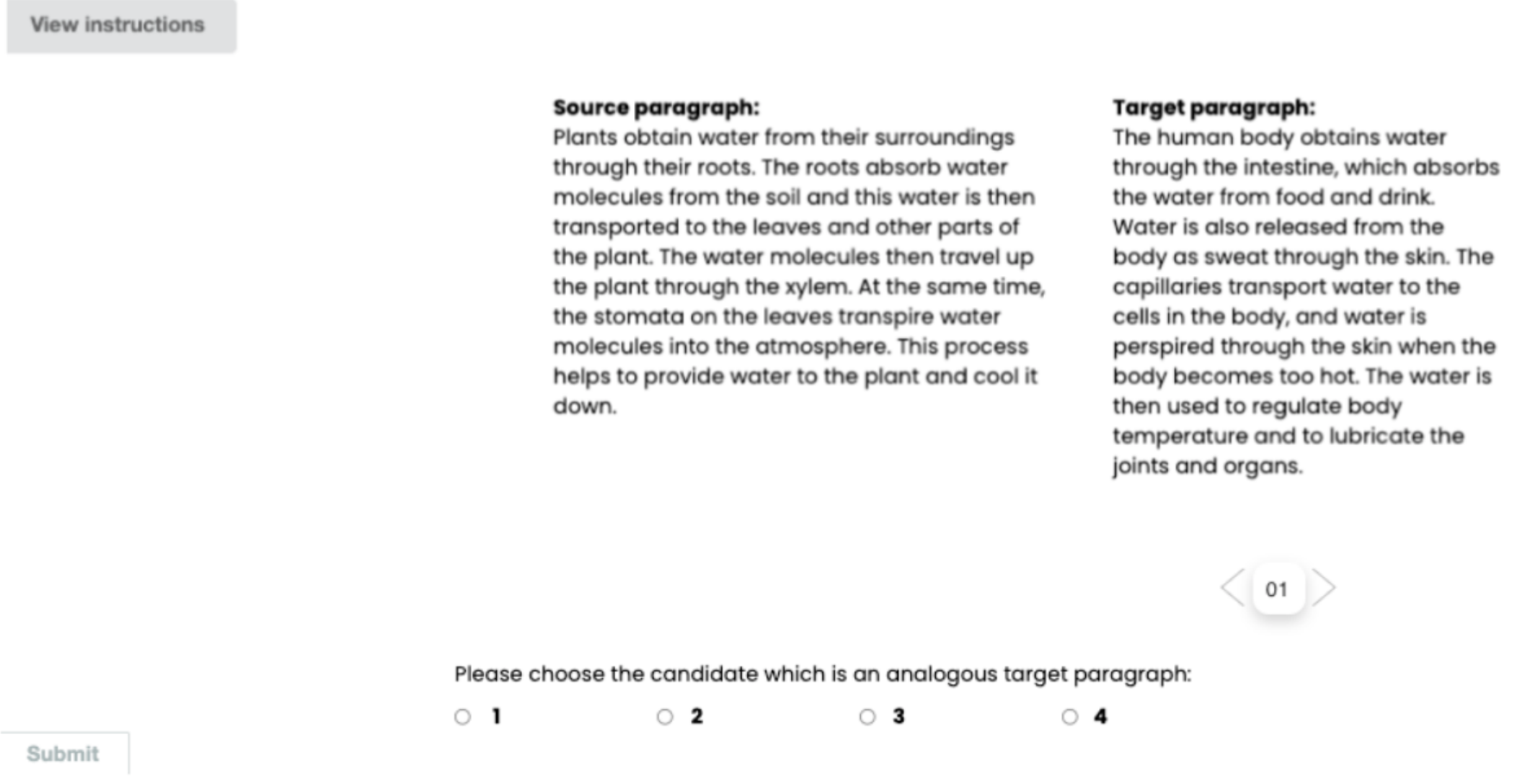}
\caption{The screen for the crowdworkers in AMT for the multiple-choice task. The annotator can press the button (right) and scroll to different target paragraphs (four in total).}
\label{fig:mturk_mc_screen}
\end{figure*}

See Figures~\ref{fig:mturk_binary_classification_screen} and \ref{fig:mturk_mc_screen} for the display screens to the crowdworkers in Amazon Mechanical Turk for the binary classification task and the multiple-choice task. See Figures~\ref{tab:binary_classification_prompt} and \ref{tab:mc_prompt} for the prompts given to both humans and models in the zero shot setting for both tasks. See Figure~\ref{tab:gpt4_few_shot_prompt} for the prompt given to GPT4 in the supervised setting. This prompt includes five mistakes as few-shot examples from the zero-shot experiment.

\subsubsection{Methods}
\label{subsubsec:appendix_methods}
In the evaluation of both the \emph{binary classification} and \emph{multiple-choice} tasks, we employ several state-of-the-art LLMs, including GPT-3.5 \cite{Brown2020LanguageMA}, GPT-4 \cite{openai2023gpt}, 
Gemini Pro \cite{team2023gemini},
FlanT5-XL (3B parameters), and FlanT5-XXL \cite{chung2022scaling}  (11B parameters). 
Other models we also considered, but we did not include are:  Falcon, Flacon-instruct \cite{penedo2023refinedweb}, and Alpaca \cite{taori2023alpaca} with their 7B version, and Vicuna \cite{vicuna2023}, LLAMA, and LLAMA2 with 7B and 13B versions \cite{touvron2023llama, touvron2023llama2}. 
We did not include the results of these models, since they failed to understand the task (chose the same candidate in the multiple-choice task or outputted an empty string).



\subsubsection{FlanT5-small Fine-tune Parameters}
\label{subsubsec:appendix_flanT5_fine_tune}
We use the default AdamW \cite{loshchilov2017decoupled} optimizer, a learning rate of 1e-5 (other learning rate values we tried are 1e-3), batch size of 16 (we tried a different batch sizes including 4, 8, and 32), and train for 7 epochs (we tried a different number of epochs in the range of 1 to 20). The metric is ``overall accuracy'' (remains relatively stable).


\begin{figure*}
\centering
\resizebox{.99\linewidth}{!}{
\begin{tabular}{p{\linewidth}} 
\toprule
\underline{\texttt{Finding analogous target subject and relations Prompt}}\\
Your task is to find an analogy between BASE and TARGET. \\
Here are the instructions for the format of relations you should provide in SIMILAR\_RELATIONS. \\
Every similar relation should be in the following format: (ENTITY1\_BASE, VERB\_BASE, ENTITY2\_BASE) \\
like (ENTITY1\_TARGET, VERB\_TARGET, ENTITY2\_TARGET). \\
ENTITY1\_BASE and ENTITY2\_BASE must be noun phrases from BASE. \\
ENTITY1\_TARGET and ENTITY2\_TARGET must be noun phrases from TARGET. \\
VERB\_BASE and VERB\_TARGET must be verbs with the same meanings. \\
\textbf{Inputs}: BASE, TARGET\_DOMAIN \\
\textbf{Outputs}: TARGET, TARGET\_FIELD, SIMILAR\_RELATIONS
\\ \\
\textbf{Inputs}: \\
\textbf{BASE}: How does the electrical circuit works? \\ 
\textbf{TARGET\_DOMAIN}: One of the fields of Engineering \\
\textbf{Outputs}: \\
\textbf{TARGET}: How does a mechanical system of water pump works? \\
\textbf{TARGET\_FIELD}: Mechanical Engineering \\
\textbf{SIMILAR\_RELATIONS}: \\
(battery, generates, electrical voltage) like (pump, generates, pressure) \\ 
(electrons, move through, copper wire) like (water, move through, pipe) \\
(resistor, decrease, voltage rate) like (valve, decrease, flow rate) \\
\bottomrule
\end{tabular}
}
\caption{A one-shot prompt for finding a target analogous subject and generating the similar relations between base and target.}
\label{tab:analogy_candidates_generation_prompt1}
\end{figure*}

\begin{figure*}
\centering
\resizebox{.99\linewidth}{!}{
\begin{tabular}{p{\linewidth}} 
\toprule
\underline{\texttt{Writing a target paragraph Prompt}}\\
Your task is to write a paragraph given SUBJECT and RELATIONS. \\
PARAGRAPH has to include RELATIONS in the text. \\ 
\textbf{Inputs}: SUBJECT, RELATIONS \\ 
\textbf{Outputs}: PARAGRAPH \\ \\
\textbf{Inputs}: \\
\textbf{SUBJECT}: How does the electrical circuit work? \\
\textbf{RELATIONS:} \\
(battery, generates, electrical voltage) \\
(electrons, move through, copper wire) \\
(resistor, decrease, voltage rate) \\
\textbf{Outputs}: \\
\textbf{PARAGRAPH}: \\
The battery generates electrical voltage. \\
This voltage creates a potential difference that causes electrons to flow through the circuit. \\
The electrical voltage causes electrons to move through the copper wire. \\ 
The electrons pass through the resistor. \\ 
The resistor presents a higher resistance to the flow of electrons, \\ 
which causes a decrease in the voltage of the circuit. \\ 
\bottomrule
\end{tabular}
}
\caption{A one-shot prompt for writing a target paragraph given the subject and the relations in target.}
\label{tab:analogy_candidates_generation_prompt2}
\end{figure*}

\begin{figure*}[t]
    \resizebox{\textwidth}{!}{%
        \fbox{\parbox[c]{9.5cm}{
            \textbf{\base: How is sediment transported across the Earth?} \\ 
            Sediment settles in a place due to gravity. \\
            The sediment breaks down into small particles. \\
            Wind or water picks up the sediment. \\
            The sediment travels along the same route as the wind or water. \\
            The sediment is deposited at new locations by the wind or water. \\
            The sediment is picked up again by new wind or water. \\ 
            The sediment travels further. \\ 
            \textbf{The sediment is deposited again in a new place.}
        }}
        \fbox{\parbox[c]{8.2cm}{
            \textbf{\targ: How is money transported across the economy?} \\ 
            Money flows through the economy. \\
            Money settles in different places. \\
            Money breaks down into smaller denominations. \\
            Investment or spending causes money to move. \\
            Money is deposited into new accounts. \\
            The money is picked up again by new investment or spending. \\
            \textbf{Money travels through the economy}. \\
            \textbf{Money is deposited again in a new place.}
        }}
    }
    \caption{An example of an analogous target paragraph (\targ) of ``How is money transported across the economy'' to a base paragraph (\base ) which is about ``How is sediment transported across the Earth?'', using one prompt for the whole task of both finding the analogous target subject and writing the paragraph, generated by GPT-3.5. 
    As we can see, using one-prompt lead to paragraphs which are mostly identical other than the nouns (``The sediment is deposited again in a new place''/``Money is deposited again in a new place''), and to artificially sounding sentences (e.g, ``Money travels through the economy'').}
\label{fig:gpt3_one_prompt_problems}
\end{figure*}

\begin{figure*}
\centering
\resizebox{.99\linewidth}{!}{
\begin{tabular}{p{\linewidth}} 
\toprule
\underline{\texttt{Analogies candidates Auto-labeling Prompt}}\\
Your task is to rate how analogous are paragraph pairs from 0 (non-analogous) to 3 (very analogous) based on whether they describe similar underlying processes or mechanisms. \\ \\ 
\textbf{SOURCE-SUBJECT}: How do floods happen?\\
\textbf{SOURCE-PARAGRAPH}: Floods happen when there is excessive rainfall which increases the water levels in rivers and streams. When the water levels get too high, the rivers and streams will overflow their banks. Additionally, heavy rainfall can also cause groundwater to rise above ground. This can lead to flooding as well.\\
\textbf{TARGET-SUBJECT}: How does a social movement develop?\\
\textbf{TARGET-PARAGRAPH}: A social movement begins with the spread of ideas among people. As more individuals learn about the movement and join it, support for the cause grows. This support often includes donations, participation in protests, and other forms of support, which helps to further the cause of the social movement.\\
\textbf{RELATIONS}: (rainfall, increases, water levels) like (ideas, spread, among people). \\ (rivers, overflow, banks) like (individuals, join, the movement). \\ (groundwater, rises, above ground) like (support, grows, for the cause)\\
\textbf{LABEL}: 0\\
\bottomrule
\end{tabular}
}
\caption{The beginning of the prompt we used for analogous paragraph's candidate auto-labeling, where no annotator classified the example as an analogy. This is one example out of 30 few-shot.
}
\label{tab:auto_label_prompt}
\end{figure*}

\begin{figure*}[t]
\centering
\includegraphics[width=.99\textwidth]{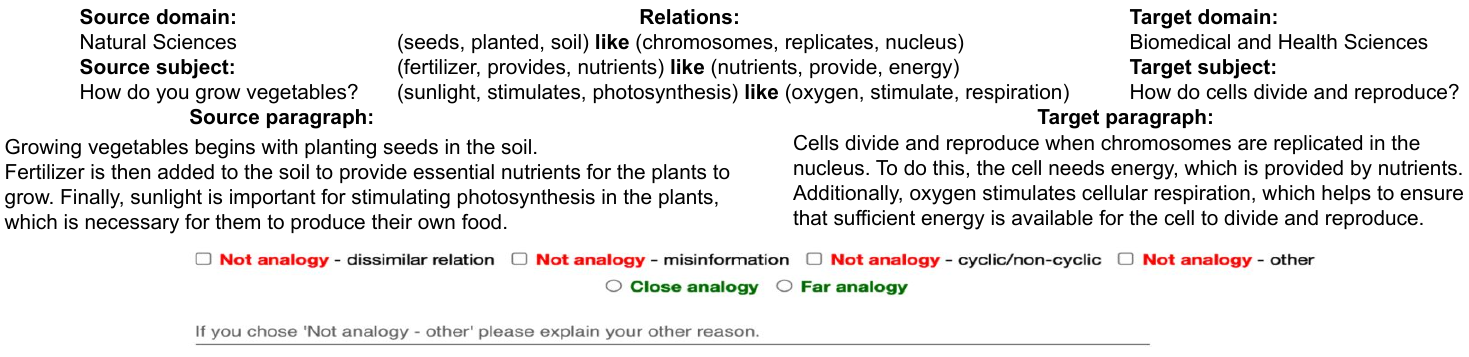}
\caption{The display screen of the annotators in Amazon Mechanical Turk. The worker has to choose one of \emph{close analogy} or \emph{far analogy} in the case of analogy, or the reasons for possible issues in the generation. If the worker chooses \emph{not analogy - other}, filling the text box with the other reason is mandatory. Note that this example is not analogy (in the current form), hence it should be postponed for further inspection. The issues raised are: \emph{dissimilar relations} (``planted'' vs. ``replicated'') and cyclic/non-cyclic (the target paragraph is a cyclic process, while the source paragraph is not)}
\label{fig:mturk_task_screen}
\end{figure*}

\begin{figure*}
\centering
\resizebox{.99\linewidth}{!}{
\begin{tabular}{p{\linewidth}} 
\toprule
\underline{\texttt{Events new order Prompt}}\\
Your task is given an INPUT\_PARAGRAPH, create a list of events in the paragraph \\
INPUT\_PARAGRAPH\_ORDER according to their order in time. 
Then, find two events where the first event must happen before the second event (a prerequisite). \\
Then, replace these events and create NEW\_PARAGRAPH\_ORDER which are the events in the new order after the replacement,
so the NEW\_PARAGRAPH\_ORDER is illogical. \\
You also need to write an EXPLANATION why NEW\_PARAGRAPH\_ORDER is illogical. \\
\textbf{Inputs}: INPUT\_PARAGRAPH \\
\textbf{Outputs}: INPUT\_PARAGRAPH\_ORDER, NEW\_PARAGRAPH\_ORDER, EXPLANATION \\ \\
\textbf{Inputs}: \\
\textbf{INPUT\_PARAGRAPH}: \\
When you get a cut on your hand, the first step is to clean the wound with antiseptic. \\
This is to help prevent infection. Then, a bandage is applied to cover the injury and protect it from further harm. \\
As time passes, the healing process begins, in which the incision is gradually closed up. \\ \\
\textbf{Outputs}: \\ 
\textbf{INPUT\_PARAGRAPH\_ORDER}: \\ 
$[1]$: A cut occurs on your hand. \\
$[2]$: The wound is cleaned with antiseptic to help prevent infection. \\
$[3]$: A bandage is applied to cover the injury and protect it from further harm. \\
$[4]$: The healing process begins and the incision is gradually closed up. \\ \\ 
\textbf{NEW\_PARAGRAPH\_ORDER}: \\
$[1]$: A cut occurs on your hand. \\
$[2]$: A bandage is applied to cover the injury and protect it from further harm. \\
$[3]$: The wound is cleaned with antiseptic to help prevent infection. \\ 
$[4]$: The healing process begins and the incision is gradually closed up. \\ \\
\textbf{EXPLANATION:} \\ 
By looking at INPUT\_PARAGRAPH\_ORDER, I found that "[2]: The wound is cleaned with antiseptic to help prevent infection." \\
must happen before "[3]: A bandage is applied to cover the injury and protect it from further harm." \\ 
because the wound should be cleaned first before a bandage can be applied. \\ 
After a bandage is applied and cover the wound, the wound cannot be cleaned, since the bandage cover the wound,
and we have no access to the wound, so it's illogical.\\
\bottomrule
\end{tabular}
}
\caption{A one-shot prompt for generating the events order in the paragraph, as well as new events order after replacing two dependent events, along with explanation of choosing the specific events.}
\label{tab:distractor_replace_events_one_shot}
\end{figure*}

\begin{figure*}
\centering
\resizebox{.99\linewidth}{!}{
\begin{tabular}{p{\linewidth}} 
\toprule
\underline{\texttt{Writing a coherent paragraph Prompt}}\\
Your task is to concatenate the EVENTS\_ORDER according to the temporal order in EVENTS\_ORDER which is wrong,
to create illogical paragraph. \\
\textbf{Inputs}: EVENTS\_ORDER \\
\textbf{Outputs}: OUTPUT\_PARAGRAPH \\ 
\\
\textbf{Inputs}: \\ 
\textbf{EVENTS\_ORDER}: \\
$[1]$: A cut occurs on your hand. \\ 
$[2]$: A bandage is applied to cover the injury and protect it from further harm. \\ 
$[3]$: The wound is cleaned with antiseptic to help prevent infection. \\ 
$[4]$: The healing process begins and the incision is gradually closed up. \\
\\
\textbf{Outputs}: \\ 
\textbf{OUTPUT\_PARAGRAPH}: \\ 
$[1]$ When you get a cut on your hand, $[2]$ the first step is to apply a bandage to cover the injury and protect it from further harm. $[3]$ After a bandage is applied, the next step is to clean the wound with antiseptic to help prevent infection.
$[4]$ As time passes, the healing process begins, in which the incision is gradually closed up.\\
\bottomrule
\end{tabular}
}
\caption{The beginning of a few-shot prompt for writing a new coherent paragraph according to the new events order after replacement of two dependent events. We show here the first out of five-shot examples in the prompt.}
\label{tab:distractor_write_paragraph_few_shot}
\end{figure*}

\begin{figure*}
\centering
\resizebox{.99\linewidth}{!}{
\begin{tabular}{p{\linewidth}} 
\toprule
\underline{\texttt{Binary Classification Task Prompt}}\\
In this task, you'll be given two paragraphs that describe scientific processes. Your goal is to decide whether the processes are analogous.
Analogy is a mapping in which the objects of one process are structurally aligned with the objects of another.
It is based on similarity of the relationships between the objects and the roles they play throughout the process, and not on the similarity between object attributes.
For example, there is an analogy between a paragraph about "How does an electrical circuit work?", and a paragraph about "How does a mechanical water pump work?". In this analogy, electrons are mapped to water: both start at some state (low voltage/low pressure), then move through something (wire/pipe), and change their state (high voltage/high pressure) because of another object (battery/pump).
Similar first order relations between the domains include:
(battery, creates, electrical voltage) like (pump, generates, pressure)
(electrons, move through, copper wire) like (water, flows through, pipe).
On the other hand, if for example the second paragraph about the pump is describing that: first the water flows inside the pipe, and following this the pump creates pressure, it changes the cause and effect relationship (higher order relation) to be different from the first paragraph about the electrical circuit, and in this case, the processes are not analogous. \\
Answer "1" if the two paragraphs describe analogous processes, and "0" if not. \\
\bottomrule
\end{tabular}
}
\caption{The prompt given for both humans and LLMs in the binary classification task}
\label{tab:binary_classification_prompt}
\end{figure*}

\begin{figure*}
\centering
\resizebox{.99\linewidth}{!}{
\begin{tabular}{p{\linewidth}} 
\toprule
\underline{\texttt{Multiple Choice Task Prompt}}\\
In this task, you'll be given a paragraph detailing a scientific process P, and four candidate paragraphs (C1, C2, C3, C4). Your goal is to identify the candiate paragraph that is analogous to P. Only one candidate paragraph is analogous to P. 
Analogy is a mapping in which the objects of one process are structurally aligned with the objects of another.
It is based on similarity of the relationships between the objects and the roles they play throughout the process, and not on the similarity between object attributes.
For example, there is an analogy between a paragraph about "How does an electrical circuit work?", and a paragraph about "How does a mechanical water pump work?". In this analogy, electrons are mapped to water: both start at some state (low voltage/low pressure), then move through something (wire/pipe), and change their state (high voltage/high pressure) because of another object (battery/pump).
Similar first order relations between the domains include:
(battery, creates, electrical voltage) like (pump, generates, pressure)
(electrons, move through, copper wire) like (water, flows through, pipe).
On the other hand, if for example the second paragraph about the pump is describing that: first the water flows inside the pipe, and following this the pump creates pressure, it changes the cause and effect relationship (higher order relation) to be different from the first paragraph about the electrical circuit, and in this case, the processes are not analogous. \\ 
Please write only the name of the candidate in your answer between C1, C2, C3, C4 that you find as describing an analogous process to the one described in P. \\
\bottomrule
\end{tabular}
}
\caption{The prompt given for both humans and LLMs in the multiple-choice task}
\label{tab:mc_prompt}
\end{figure*}

\begin{figure*}
\centering
\resizebox{.79\linewidth}{!}{
\begin{tabular}{p{\linewidth}} 
\toprule
\underline{\texttt{GPT4 (few-shot) Binary Task Prompt}}\\
In this task, you'll be given two paragraphs that describe scientific processes. Your goal is to decide whether the processes are analogous.
Analogy is a mapping in which the objects of one process are structurally aligned with the objects of another.
It is based on similarity of the relationships between the objects and the roles they play throughout the process, and not on the similarity between object attributes.
For example, there is an analogy between a paragraph about "How does an electrical circuit work?", and a paragraph about "How does a mechanical water pump work?". In this analogy, electrons are mapped to water: both start at some state (low voltage/low pressure), then move through something (wire/pipe), and change their state (high voltage/high pressure) because of another object (battery/pump).
Similar first order relations between the domains include:
(battery, creates, electrical voltage) like (pump, generates, pressure)
(electrons, move through, copper wire) like (water, flows through, pipe).
On the other hand, if for example the second paragraph about the pump is describing that: first the water flows inside the pipe, and following this the pump creates pressure, it changes the cause and effect relationship (higher order relation) to be different from the first paragraph about the electrical circuit, and in this case, the processes are not analogous.
Answer "1" if the two paragraphs describe analogous processes, and "0" if not. \\
\textbf{Inputs:} First Paragraph, Second Paragraph \\
\textbf{Outputs:} Answer \\

\textbf{First Paragraph:} \\
A wind-powered power station generates electricity by using wind turbines that capture kinetic energy from the wind.
This energy is then converted by a generator into electricity, which then flows through power lines to be used in homes and businesses.
The wind turbine captures the kinetic energy of the wind and converts it into electrical energy by spinning a generator, which then causes electricity to flow through the power lines. \\
\textbf{Second Paragraph:} \\
Solar energy is captured by the solar panels. The electricity generated can then be used to power various electrical appliances. Afterward, the generator converts solar energy into electricity. Finally, electricity flows through wires to reach the appliances. \\
\textbf{Answer:} 0 \\

\textbf{First Paragraph:} \\
Floods happen when heavy rain saturates the soil, causing water to accumulate in low-lying areas.
The excess water can cause the ground to become unstable, leading to flooding. \\
\textbf{Second Paragraph:} \\
A heavy snowfall saturates the mountain slope. This instability then causes the snow to break loose. After the snow breaks loose, it accumulates on the steep slopes. As the snow accumulates, it becomes increasingly unstable. Finally, the avalanche is created. \\ 
\textbf{Answer:} 0 \\ 

\textbf{First Paragraph:} \\
Bats use echolocation to navigate and find food. They emit high frequency sound waves that bounce off of objects in their environment. The bats then receive the echoes and interpret the information to locate their prey and navigate their surroundings. The echo provides the bats with information about the shape, size, and distance of the object. \\ 
\textbf{Second Paragraph:} \\
Submarines interpret the echo to determine the distance and size of the object. After interpreting the echo, they emit sound waves, which travel through the water and bounce off the objects. These sound waves are then received back as an echo. Finally, submarines use sonar technology to detect objects in the water. \\
\textbf{Answer:} 0 \\

\textbf{First Paragraph:} \\
Floods happen when there is an excessive amount of rainfall in a certain area.
The rain causes the ground to be saturated, leading to flooding.
The flood water can damage buildings and crops, as well as cause disruption to transport and other infrastructure.
In addition, rivers can overflow their banks due to the high levels of water, leading to even further flooding. \\ 
\textbf{Second Paragraph:} \\
The wind causes vibration and can damage structures, so engineers must design bridges to withstand the forces the wind exerts.
The wind can produce a force that pushes the bridge sideways and could cause it to collapse if not designed properly.
Engineers must build bridges in such a way that the wind does not exert too much force on the bridge, and that the bridge is able to withstand the vibration caused by the wind." \\ 
\textbf{Answer:} 1 \\ 

\textbf{First Paragraph:} \\
Igneous rocks are formed from molten material. This molten material is known as magma and it solidifies into rock as it cools. As the magma cools, crystals form within it, creating the igneous rock. The combination of the cooling of magma and the formation of crystals is what creates igneous rock. \\ 
\textbf{Second Paragraph:} \\
People come together to form a social movement.
The organization of people, who have a common goal, creates a movement.
Social movements are formed from collective action, as individuals come together to fight for a shared cause.
By uniting, people can accomplish goals that they cannot achieve on their own. \\ 
\textbf{Answer:} 1 \\
\bottomrule
\end{tabular}
}
\caption{The few-shot prompt given to GPT4 on the binary classification task. It includes 5 examples of mistakes made by GPT4 in the zero-shot experiment. Three on distractors, one on analogy, and one on random.}
\label{tab:gpt4_few_shot_prompt}
\end{figure*}

\label{sec:appendix}

\end{document}